\documentclass[letterpaper]{article} % DO NOT CHANGE THIS
\usepackage{aaai25}  % DO NOT CHANGE THIS
\usepackage{times}  % DO NOT CHANGE THIS
\usepackage{helvet}  % DO NOT CHANGE THIS
\usepackage{courier}  % DO NOT CHANGE THIS
\usepackage[hyphens]{url}  % DO NOT CHANGE THIS
\usepackage{graphicx} % DO NOT CHANGE THIS
\usepackage{natbib}  % DO NOT CHANGE THIS AND DO NOT ADD ANY OPTIONS TO IT
\usepackage{caption} % DO NOT CHANGE THIS AND DO NOT ADD ANY OPTIONS TO IT
\usepackage{booktabs}
\usepackage{makecell}
\usepackage{amsmath}
\usepackage{amsfonts}
\usepackage{multirow}
\usepackage{algorithm}
\usepackage{algorithmic}

\usepackage{newfloat}
\usepackage{listings}
\DeclareCaptionStyle{ruled}{labelfont=normalfont,labelsep=colon,strut=off} % DO NOT CHANGE THIS
\floatstyle{ruled}
\newfloat{listing}{tb}{lst}{}
\floatname{listing}{Listing}
\newcommand{\ie}{\textit{i.e.}}
\newcommand{\eg}{\textit{e.g.}}
\newcommand{\ours}{{SPRING}}
\newcommand{\ttt}{\texttt}

\title{One Token Can Help! Learning Scalable and Pluggable Virtual Tokens \\ for Retrieval-Augmented Large Language Models}
\author{
    Yutao Zhu,
    Zhaoheng Huang,
    Zhicheng Dou\thanks{Corresponding author.}, 
    Ji-Rong Wen \\
}
\affiliations{
    Gaoling School of Artificial Intelligence, Renmin University of China \\
    yutaozhu94@gmail.com, \{huangzh, dou, jrwen\}@ruc.edu.cn
}

\begin{document}

\maketitle

\begin{abstract}
Retrieval-augmented generation (RAG) is a promising way to improve large language models (LLMs) for generating more factual, accurate, and up-to-date content. Existing methods either optimize prompts to guide LLMs in leveraging retrieved information or directly fine-tune LLMs to adapt to RAG scenarios. Although fine-tuning can yield better performance, it often compromises the LLMs' general generation capabilities by modifying their parameters. This limitation poses challenges in practical applications, especially when LLMs are already deployed, as parameter adjustments may affect their original functionality. To address this, we propose a novel method that involves learning scalable and pluggable virtual tokens for RAG. By maintaining the LLMs' original parameters and fine-tuning only the embeddings of these pluggable tokens, our approach not only enhances LLMs' performance but also preserves their general generation capabilities. Furthermore, we design several training strategies to improve the scalability, flexibility, and generalizability of our method. Comprehensive experiments across 12 question-answering tasks demonstrate the superiority of our approach.
\end{abstract}


\section{Introduction}
Large language models (LLMs) have achieved remarkable performance across various natural language processing tasks~\cite{gpt-3,gpt-4,llama}. Despite their extensive parameters enabling them to learn rich knowledge during pre-training, LLMs may still generate hallucinated, outdated, or inaccurate content, especially in scenarios requiring long-tail knowledge~\cite{DBLP:journals/csur/JiLFYSXIBMF23,DBLP:journals/corr/abs-2309-01219}.
To address this problem, retrieval-augmented generation (RAG) has emerged as a pivotal strategy. By explicitly decoupling knowledge retrieval from the backbone LLMs, such architectures have achieved more accurate and reliable content generation and shown particularly enhanced performance on knowledge-intensive tasks such as open-domain question answering~\cite{kilt,slimplm,bider}.
% ~\cite{DBLP:conf/nips/LewisPPPKGKLYR020,kilt,slimplm,bider}. 

\begin{figure*}
    \centering
    \includegraphics[width=.85\linewidth]{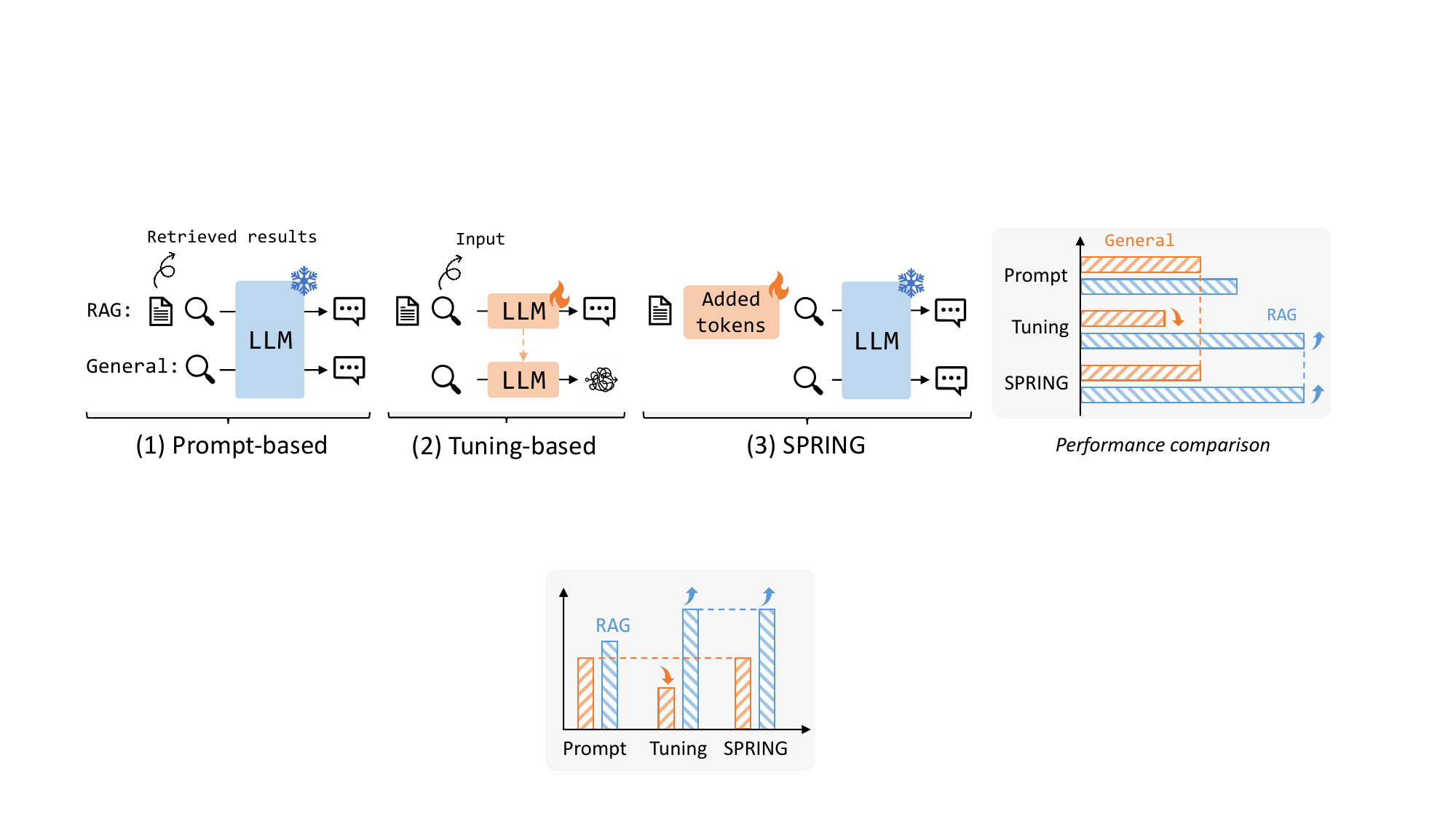}
    \caption{Illustration of existing methods for RAG and our proposed method. 
    Our method can improve LLMs' performance in RAG scenarios by incorporating trainable virtual tokens, and these tokens can be removed to preserve the general generation abilities in non-RAG scenarios.}
    \label{fig:intro}
    %\vspace{-5px}
\end{figure*}

Existing efforts in RAG development can be roughly categorized into two groups (as illustrated in Figure~\ref{fig:intro}). The first group leverages the in-context learning capabilities of LLMs by incorporating retrieved information into the input along with appropriate prompts~\cite{replug,incontext-ralm}. This allows for straightforward application to any \textit{off-the-shelf} LLM without tuning its parameters. However, its effectiveness largely depends on the human experience in crafting effective prompts and the LLM's ability to interpret these prompts. The second group focuses on training LLMs to enhance their performance in RAG scenarios. This training might involve either \textit{end-to-end pre-training}~\cite{realm,retro} or \textit{fine-tuning}~\cite{radit,filco} for specific tasks. These approaches can often lead to better performance, but they require significant computational resources. Recently, parameter-efficient fine-tuning techniques, such as LoRA~\cite{lora}, have been widely studied, significantly reducing training costs. These methods can optimize the LLMs' parameters for RAG, but unfortunately compromise the model's general abilities in non-RAG scenarios, such as commonsense reasoning and in-context learning. All these limitations prevent their application to LLMs already operational in real-world settings.

Therefore, a critical research problem arises: \textit{Is it possible to enhance LLMs' performance under RAG scenarios while preserving their general generation capabilities?} To achieve this, we introduce a novel, lightweight tuning method named \textbf{\ours{}}, which learns \underline{\textbf{S}}calable and \underline{\textbf{P}}luggable vi\underline{\textbf{R}}tual tokens for retr\underline{\textbf{I}}eval-augme\underline{\textbf{N}}ted \textbf{G}eneration. Our basic idea is to add trainable virtual tokens to help LLMs learn RAG problems. Through fine-tuning, these virtual tokens effectively enhance the LLM's capability to understand retrieved information and its correlation with user inputs. Importantly, as the LLM's original parameters are frozen, its general generation abilities are preserved without any loss. During inference, when retrieval is triggered, these trained virtual tokens can be simply added to the prompt, which includes both the retrieved results and user input, thereby significantly enhancing performance. Moreover, we employ a scalable training approach, allowing the number of virtual tokens to be adjusted according to the needs of the inference scenario. Various training strategies have been implemented to further improve the generalizability of our method, ensuring robustness regardless of the number of the retrieved results.

In experiments, \ours{} is trained with the base and instruction fine-tuned versions of Mistral-7b, LLaMA-2-7b, and LLaMA-2-13b models and evaluated on 12 commonly used QA datasets, covering both in-domain and out-of-domain scenarios. The experimental results demonstrate that \ours{} not only effectively improves the RAG performance of LLMs but also successfully preserves their general generation capabilities. Overall, the \ours{} method exhibits four main characteristics:

$\bullet$ \textbf{Lightweight yet effective}. Instead of updating the full parameters of the LLMs, we opt to freeze the pre-trained models and only learn the embeddings for the added virtual tokens. For example, adding 50 tokens to the Mistral-7b model introduces only 0.2M parameters in total. Despite these minimal parameters, \ours{} improves the average EM and F1 scores by more than 43\% and 17\% across 12 QA datasets, respectively.

$\bullet$ \textbf{Scalable}. With our proposed scalable training approach, \ours{} can be effective with any number of virtual tokens ($k\in[1, 50]$ in our experiments). Remarkably, even just one token can substantially improve the LLMs' performance in RAG scenarios.

$\bullet$ \textbf{Pluggable}. Owing to its lightweight design, \ours{} can be applied in a plug-and-play manner. When retrieval is triggered, simply adding the virtual tokens can lead to better performance. In non-RAG scenarios, the virtual tokens are not added so the LLMs' original capabilities can be well preserved. This characteristic is crucial for LLMs that have already been deployed for practical use.

$\bullet$ \textbf{Generalizable}. Our robust training strategies ensure that \ours{} is adaptable to different retrievers and various numbers of retrieved results. Consequently, there is no need to retrain \ours{} with each update to the retrieval system, enhancing its practicality and efficiency.

\section{Related Work}

% ~\cite{DBLP:conf/nips/LewisPPPKGKLYR020,realm,llm4ir}
% ~\cite{replug,incontext-ralm,retro,radit,inters}

\paragraph{Retrieval-Augmented Generation} Compared to standard text generation, retrieval-augmented generation (RAG) incorporates a retrieval module that accesses external knowledge to enhance generation quality~\cite{DBLP:conf/nips/LewisPPPKGKLYR020,realm,llm4ir,flashrag}. The mainstream RAG follows a ``retrieve-then-read'' paradigm, where the retrieval module provides external knowledge as additional context, which is then read by generation models to produce the final output~\cite{replug,incontext-ralm,retro,radit,inters}. To optimize the use of external knowledge, some methods focus on crafting effective prompts that guide the utilization of retrieved information~\cite{replug,incontext-ralm}. These prompt-based methods are applicable to any LLM without tuning its parameters. However, they depend heavily on skillful prompt writing and the LLMs' ability to understand instructions. In contrast, other studies attempts to directly train the model to better use the retrieved knowledge. For example, REALM~\cite{realm} and RETRO~\cite{retro} incorporate retrieval in end-to-end retrieval-augmented pre-training. RA-DIT~\cite{radit} employs fine-tuning to enhance LLMs' retrieval understanding. These tuning-based methods often yield better performance than prompt-based methods by optimizing model parameters for RAG. However, they may compromise the LLMs' general capabilities, particularly in non-retrieval scenarios. Different from existing methods, we design a new lightweight tuning method for RAG. It is a plug-and-play module that enhances RAG performance using trainable virtual tokens, which can be removed in non-RAG scenarios to preserve the LLMs' general generation abilities.

\paragraph{Parameter-Efficient Fine-Tuning} The paradigms of ``pre-training then fine-tuning'' have demonstrated efficacy across various natural language~\cite{bert,t5,gpt-2} and vision tasks~\cite{moco,vit,simclr}. The common fine-tuning process involves tuning all parameters of a model, which is computational intensive, especially for LLMs. To address this, parameter-efficient fine-tuning (PEFT)~\cite{peft} approaches have been developed. These approaches freeze most of the pre-trained models' parameters, yet still manage to achieve comparable performance on downstream tasks. PEFT has been widely studied~\cite{peft_survey}, and typical methods including adapter-based tuning~\cite{DBLP:conf/icml/HoulsbyGJMLGAG19,DBLP:conf/emnlp/LinMF20,DBLP:conf/nips/RebuffiBV17,DBLP:conf/iclr/ChenDWHLDQ23}, low-rank adaptation (LoRA)~\cite{lora,qlora}, and prompt tuning~\cite{prefix-tuning,prompt_tuning,p_tuning,p_tuning_v2,DBLP:conf/naacl/QinE21}. Adapter-based tuning inserts lightweight modules into a model's existing layers and have been extended to various domains~\cite{DBLP:journals/ijcv/GaoGZMFZLQ24,llm_adapter,llama_adapter}. LoRA~\cite{lora} introduces trainable low-rank matrices that adjust the model’s weight updates, achieving promising fine-tuning performance on LLMs~\cite{llm_adapter}. Prompt tuning incorporates a series of trainable prompt tokens to LLMs. These tokens can be inserted either to the input layer only~\cite{prompt_tuning,p_tuning} or to all of the intermediate layers~\cite{prefix-tuning,p_tuning_v2}. In this paper, we proposes a novel prompt tuning method, {\ours{}}, specifically designed for RAG scenarios. Our method introduces virtual tokens between retrieved results and the input, exploiting the auto-regressive generation paradigm to improve the model's ability to utilize retrieved information. Additionally, it is designed to be scalable and pluggable, thus broadening its application scope while preserving the original generative capabilities of LLMs.
% Different from existing approaches, our method adds virtual tokens between retrieved results and input to leverage the characteristics of auto-regression generation paradigm so as to enhance the LLMs' ability of using the retrieved information. Besides, it is designed as scalable and pluggable to extend its application scope while keeping the original capabilities of the LLMs without damage.

\begin{figure*}
    \centering
    \includegraphics[width=.85\linewidth]{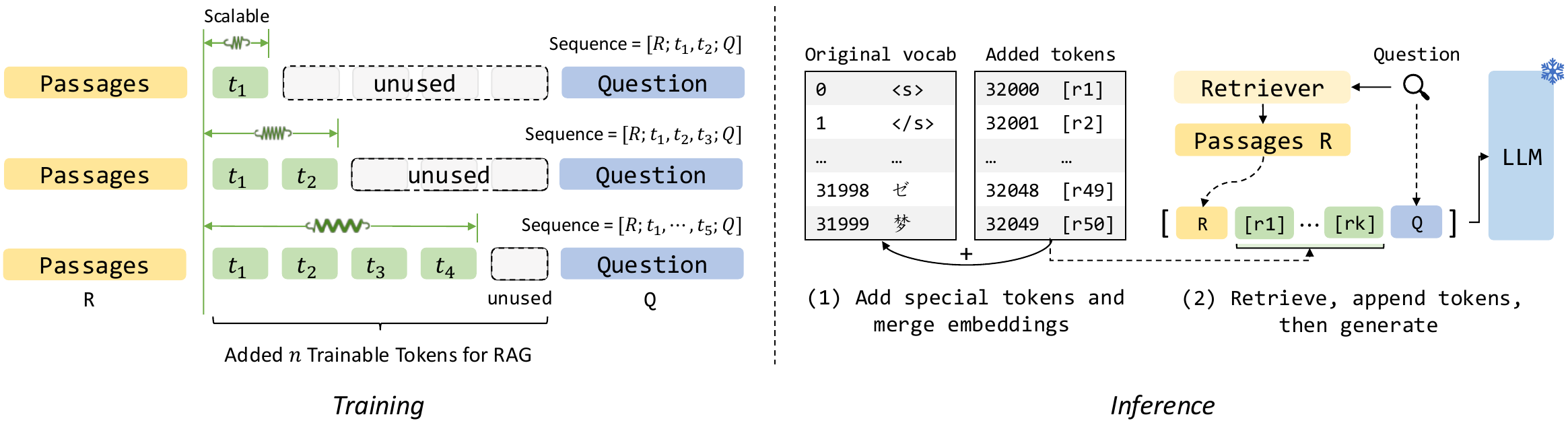}
    \caption{Illustration of \ours{}. Only the embeddings of the added $n$ tokens are trainable during fine-tuning. The added tokens are scalable where any first $k (k\leq n)$ tokens can be used in inference.}
    \label{fig:method}
    %\vspace{-5px}
\end{figure*}

\section{Methodology}
To take advantage of both the flexibility of prompt-based methods and the efficacy of fine-tuning-based methods, we propose \ours{} to learn scalable and pluggable virtual tokens for retrieval-augmented generation (RAG). 
% In this section, we will first briefly formulate the problem of RAG~(\S\ref{sec:formulation}), and then dive into the details of our method~(\S\ref{sec:method}). 

\subsection{Problem Formulation}\label{sec:formulation}
Language models are designed to calculate the probability distribution over sequences of natural language texts. Auto-regressive models are commonly used for this through next-token prediction:
\begin{align}
    p_{\text{LM}} = \prod_{i=1}^{m}p_{\theta}(x_i|x_{<i}),
\end{align}
where $x_{<i}$ denotes the sequence of tokens preceding $x_i$ at each step, and $\theta$ represents the parameters of the model. For RAG, a retrieval corpus $\mathcal{D}$ and a retriever ${M}$ are introduced. Then, the generation process is conditioned on both $x_{<i}$ and the retrieved results ${R}={M}_\mathcal{D}(x_{<i})$ as:
\begin{align}
    p_{\text{RAG}} &= \prod_{i=1}^{m}p_{\theta}(x_i|{R};x_{<i}), \\
    p_{\text{RAG-QA}} &= \prod_{i=1}^{m}p_{\theta}(a_i|{R};Q;a_{<i}). \label{eq:ralm}
\end{align}
Note that here $x_{<i}$ serves as the \textit{query} for retrieval. In question-answering (QA) tasks, $x_{<i}$ is usually the question $Q$, and the learning objective is to generate the right answer $A=\{a_i\}_{i=1}^{m}$. The retriever can yield multiple passages, which can be concatenated as a long text sequence using proper separator such as ``$\backslash$n$\backslash$n''. For brevity, this formulation directly concatenates the retrieved results ${R}$ with the question $Q$, omitting more complex prompt designs. Henceforth, we will use the notations in QA tasks as our evaluation is performed on them.

\subsection{Scalable and Pluggable Virtual Tokens for RAG}\label{sec:method}
Our \ours{} method, shown in the left side of Figure~\ref{fig:method}, introduces trainable virtual tokens into the input to optimize LLMs for RAG scenarios. Specifically, following the notation in Equation~(\ref{eq:ralm}), we add $n$ trainable tokens $T=[t_1,t_2,\cdots,t_n]$ \textit{between} the retrieved results ${R}$ and the input $Q$. The generation process can then be described as:
\begin{align}
    p_{\text{\ours{}}} = \prod_{i=1}^{m}p_{\theta,\delta}(a_i|{R};[t_1,t_2,\cdots,t_n];Q;a_{<i}), \notag
\end{align}
where $\delta\in\mathbb{R}^{n\times d}$ represents the added parameters of the trainable tokens (\ie, their embeddings), and $d$ is the embedding size of the LLM. $\theta$ denotes the parameters of the backbone LLM, which are \emph{frozen} during training. Given that $|\delta| \ll |\theta|$, our method is \emph{highly efficient} for training. For example, with the Mistral-7b model (where $d=4,096$), when $n=50$ tokens are added, we only add and train $50\times 4,096=0.2\text{M}$ parameters, approximately 0.003\% of the full model. 

% [@TO-DO]: To improve the 

Importantly, we place the virtual tokens $T$ between the retrieved results $R$ and the question $Q$ for two main reasons: (1) In the auto-regressive generation paradigm, positioning the tokens after the retrieved results allows them to attend to this information, thereby aiding the model's comprehension. (2) Recent studies have indicated that LLMs are particularly sensitive to the end of an input~\cite{lost_in_middle}. By consistently placing these virtual tokens before the question across all test samples, we aim to mitigate any potential adverse effects on the understanding of the question.

\paragraph{Scalable} In practical developments, LLMs are often constrained by their maximum input lengths, limiting the number of tokens available for retrieval augmentation (especially when the retrieved results are very long). Therefore, it is desired to design a mechanism so that any number of virtual tokens can be used in the inference to improve RAG performance. To achieve this, we propose an optimization strategy working like a ``spring'' (as shown in Figure~\ref{fig:method}). Specifically, for a given sample $\{R,Q,A\}$ with the total number of added tokens $n$, we randomly select a number $k (k \leq n)$ and utilize the \textit{first} $k$ virtual tokens $t_{1:k}$ to construct the training example as $[R;t_1,t_2,\cdots,t_k;Q]$. This method allows for the flexible optimization of any number of virtual tokens. Consequently, the number of virtual tokens incorporated during inference can be arbitrarily selected based on the requirements of the application. The effectiveness of this strategy and its comparison with other methods are further discussed in our experiments.

\paragraph{Pluggable} Due to its designed structure, our method provides considerable flexibility in application. Practically, if user input is assessed to require external knowledge, our virtual tokens can be simply appended after the retrieval results and then fed, along with the user input, into the LLM for generation. In contrast, if the user input does not necessitate retrieval, it can be processed directly by the LLM. As our approach does not adjust the original parameters of the LLM, it preserves the model’s inherent capabilities. This feature is particularly important for industry or business since existing LLMs may have already been deployed for multiple purposes; our method enhances the retrieval understanding capabilities of these models without compromising their existing functionalities.

% \begin{figure}[t]
%     \centering
%     \includegraphics[width=\linewidth]{figures/inference.pdf}
%     \caption{Instructions for using \ours{}.}
%     \label{fig:inference}
% \end{figure}

\paragraph{Inference} We illustrate the instructions for using our \ours{} in the right side of Figure~\ref{fig:method}. After training, the embeddings of the added tokens have been optimized for RAG, but these tokens do not correspond to any existing tokens in the vocabulary. To make them easy to use, we can add some special tokens (\eg, \ttt{[r1]}, $\cdots$, \ttt{[r50]}) to the vocabulary and initialize their embeddings with the trained embeddings. Then, during inference, after obtaining the retrieved results $R$, we can add any number of these special tokens (\eg, \ttt{[r1]} $\cdots$ \ttt{[rk]}) after $R$ and input them with the question $Q$ to the LLMs for generation. We also provide an example code snippet in Appendix for using our method in practice.

We refer to our method as \ours{} due to its scalable and pluggable nature, making it particularly well-suited for enhancing existing LLMs that have already been deployed. Additionally, it effectively bridges the gap between retrieved results and user input, significantly improving the LLMs' capabilities in understanding the retrieved external knowledge.

\begin{table*}[t]
    \centering
    \small
    \setlength{\tabcolsep}{1.2mm}{
    \begin{tabular}{llcccccccccccc}
    \toprule
        &  & \multicolumn{6}{c}{with Retrieval} & \multicolumn{6}{c}{without Retrieval} \\
        \cmidrule(r){3-8}\cmidrule(l){9-14}
        Dataset & Metric & Concat & Prompt & Prefix & LoRA & \ours{} & \ours{}$^+$ & Concat & Prompt & Prefix & LoRA & \ours{} & \ours{}$^+$ \\
    \midrule
        \multicolumn{2}{l}{\textit{Tuning Parameters}} & 0 & 0 & 0.2M & 4M & 0.2M & 4.2M & 0 & 0 & 0.2M & 4M & 0.2M & 4.2M \\
    \midrule
        \multirow{2}{*}{Trivia QA} & EM & 0.00 & 57.79 & 11.74 & 62.76 & \textbf{65.71} & 63.89 & 0.01 & 39.90 & 0.00 & 0.03 & \textbf{46.56} & 43.37 \\
         & F1 & 65.60 & 80.33 & 59.97 & 85.44 & 85.26 & \textbf{85.94} & 63.96 & 69.72 & 28.30 & 34.91 & 74.48 & \textbf{74.89} \\
        \multirow{2}{*}{NQ} & EM & 0.00 & 28.99 & 13.04 & 47.95 & 42.35 & \textbf{49.78} & 0.00 & 13.36 & 0.00 & 0.00 & 18.80 & \textbf{25.54} \\
         & F1 & 41.77 & 58.72 & 38.22 & 74.15 & 70.73 & \textbf{75.22} & 43.74 & 48.63 & 17.74 & 29.10 & 55.75 & \textbf{60.82} \\
        \multirow{2}{*}{HQA} & EM & 0.00 & 26.36 & 5.79 & 39.95 & 35.26 & \textbf{41.14} & 0.00 & 17.07 & 0.00 & 0.03 & 20.15 & \textbf{23.38} \\
         & F1 & 44.91 & 56.15 & 42.59 & 68.93 & 65.44 & \textbf{69.95} & 47.54 & 49.50 & 17.45 & 27.57 & 54.79 & \textbf{57.99} \\
        \multirow{2}{*}{SQuAD} & EM & 0.00 & 23.92 & 7.19 & 35.71 & 33.67 & \textbf{35.98} & 0.00 & 8.61 & 0.00 & 0.00 & 12.71 & \textbf{13.90} \\
         & F1 & 43.05 & 57.66 & 39.75 & 68.05 & 66.99 & \textbf{68.41} & 43.32 & 46.81 & 21.88 & 27.51 & 53.58 & \textbf{54.93} \\
        \multirow{2}{*}{WebQ} & EM & 0.00 & 17.53 & 4.44 & 43.65 & 31.84 & \textbf{48.10} & 0.00 & 14.79 & 0.00 & 0.00 & 24.95 & \textbf{28.81} \\
         & F1 & 37.46 & 52.10 & 31.55 & 71.99 & 64.78 & \textbf{73.88} & 44.34 & 50.60 & 20.14 & 32.36 & 59.83 & \textbf{62.95} \\
        \multirow{2}{*}{2Wiki} & EM & 0.00 & 22.64 & 4.38 & 35.93 & 31.80 & \textbf{37.12} & 0.00 & 23.45 & 0.00 & 0.01 & 24.62 & \textbf{28.60} \\
         & F1 & 47.77 & 55.58 & 41.82 & 63.85 & 62.03 & \textbf{64.60} & 52.83 & 53.55 & 21.14 & 37.09 & 56.83 & \textbf{59.12} \\
        \multirow{2}{*}{CoQA} & EM & 0.00 & 8.20 & 1.56 & 12.89 & 13.28 & \textbf{13.87} & 0.00 & 8.59 & 0.00 & 0.00 & 9.96 & \textbf{12.50} \\
         & F1 & 27.98 & 36.72 & 20.02 & 41.19 & \textbf{42.41} & 42.04 & 32.97 & 36.58 & 13.15 & 18.99 & 39.96 & \textbf{41.36} \\
        \multirow{2}{*}{MS MARCO} & EM & 0.00 & 5.73 & 0.60 & 8.13 & 6.57 & \textbf{8.27} & 0.00 & \textbf{2.56} & 0.00 & 0.01 & 2.09 & \textbf{3.24} \\
         & F1 & 56.44 & 53.81 & 50.56 & 54.81 & 53.48 & \textbf{55.90} & 49.50 & 47.75 & 47.44 & 52.44 & \textbf{51.41} & 49.84 \\
        \multirow{2}{*}{PopQA*} & EM & 0.00 & 39.79 & 10.02 & 47.15 & \textbf{48.71} & 46.98 & 0.00 & 16.05 & 0.00 & 0.00 & \textbf{20.25} & 18.70 \\
         & F1 & 56.49 & 68.26 & 44.54 & 73.12 & \textbf{73.90} & 73.29 & 53.61 & 54.85 & 20.39 & 25.09 & \textbf{58.32} & 58.05 \\
        \multirow{2}{*}{Fermi*} & EM & 0.00 & 0.06 & 0.00 & 0.12 & 0.18 & \textbf{0.24} & 0.00 & 0.06 & 0.06 & 0.00 & 0.06 & 0.19 \\
         & F1 & 21.66 & 18.16 & 12.65 & 29.63 & \textbf{31.15} & 30.67 & 25.51 & 17.84 & 20.33 & 25.42 & 29.28 & 30.37 \\
        \multirow{2}{*}{Musique*} & EM & 4.01 & 4.05 & 0.04 & 10.86 & 8.80 & \textbf{12.58} & 0.00 & 1.93 & 0.66 & 0.17 & 3.64 & \textbf{4.43} \\
         & F1 & 38.79 & 38.64 & 20.91 & 52.65 & 48.74 & \textbf{53.65} & 42.30 & 36.18 & 35.22 & 45.31 & 44.93 & \textbf{47.70} \\
        \multirow{2}{*}{Bamboogle*} & EM & 12.80 & 12.80 & 0.00 & 24.22 & 22.66 & \textbf{28.13} & 0.00 & 4.69 & 3.20 & 0.00 & 12.00 & \textbf{12.80} \\
         & F1 & 45.81 & 48.13 & 13.67 & 59.05 & 56.95 & \textbf{62.00} & 42.61 & 42.24 & 36.22 & 45.14 & \textbf{47.69} & 46.89 \\
    \midrule
        \multirow{2}{*}{Average} & EM & 1.40 & 19.80 & 4.90 & 30.78 & 28.40 & \textbf{32.17} & 0.00 & 12.59 & 0.33 & 0.02 & 16.32 & \textbf{17.96} \\
         & F1 & 43.98 & 51.30 & 34.69 & 61.91 & 60.15 & \textbf{62.96} & 45.19 & 46.19 & 24.95 & 33.41 & 52.24 & \textbf{53.74} \\
    \bottomrule
    \end{tabular}
    }
    \caption{Evaluation results of different methods on twelve QA datasets. The retriever is \texttt{E5-large} model, and the number of retrieved passages is set as three. The number of virtual tokens used in \ours{} is set as 50. $^*$PopQA, Fermi, Musique, and Bamboogle are invisible during training. ``Prefix'' stands for prefix-tuning, and ``SPRING$^{+}$'' is trained based on the LoRA's checkpoint. The best results are in \textbf{bold}.}
    %\vspace{-5px}
    \label{tab:in-domain}
\end{table*}

\section{Experiment}

\subsection{Datasets and Retrievers}
We conduct experiments on twelve commonly used question-answering datasets, including TriviaQA (TQA)~\cite{triviaqa}, Natural Questions (NQ)~\cite{nq}, HotpotQA (HQA)~\cite{HotpotQA}, SQuAD 1.1~\cite{squad}, Web Questions (WebQ)~\cite{web_questions}, 2WikiMultiHopQA (2Wiki)~\cite{2wikimultihopqa}, CoQA~\cite{coqa}, MS MARCO~\cite{ms_marco}, PopQA~\cite{popqa}, Fermi~\cite{fermi}, Musique~\cite{musique}, and Bamboogle~\cite{bamboogle}. These datasets are publicly available at HuggingFace or their official websites. To evaluate the generalizability of the methods, we select PopQA, Fermi, Musique, and Bamboogle as held-out datasets. We mix the training set of all remaining datasets for training. For all datasets, we prioritize the use of test sets for evaluation purposes. In cases where the test set is not available, we utilize the development set instead. It is worth noting that, though some datasets have provided golden reference passages for the answer, we do not use them in our experiment but use the passages retrieved from the retrieval sets in both training and inference stages, which aligns with practical applications. Exact match (EM) and F1 score are employed as evaluation metrics.

For the retrieval sets, we follow previous studies~\cite{DBLP:journals/corr/abs-2310-01558} and use the combination of Wikipedia and MS MARCO datasets as the retrieval corpus. Wikipedia contains high-quality human knowledge, which is helpful for many knowledge-intensive tasks. MS MARCO contains a large amount of web pages, which can provide information necessary for curating some natural language questions. We use the datasets that have already been preprossed into passages and released on HuggingFace.\footnote{Wikipedia passages: \url{https://huggingface.co/datasets/Tevatron/wikipedia-nq-corpus}. MS MARCO passages: \url{https://huggingface.co/datasets/Tevatron/msmarco-passage-corpus}.} The Wikipedia set has 21M passages, while the MS MARCO set has 8M passages. More details are provided in Appendix.

We use \texttt{E5-large}~\cite{e5} as the main retriever in our experiments. The impact of other retrievers, \ie, \texttt{BM25}~\cite{bm25}, \texttt{BGE-base}~\cite{bge}, and \texttt{E5-base}, is studied in our further analysis. Among these retrievers, \texttt{BM25} is a non-neural sparse retrieval algorithm, while others are neural-based dense retrievers. In general, dense retrievers perform better on several benchmarks~\cite{mteb}. 

\subsection{Baseline Methods}
We consider both the base and instruction fine-tuned versions of {Mistral-7b}, {LLaMA-2-7b}, and {LLaMA-2-13b} as the backbone models, and compare our \ours{} with the following baselines. 

$\bullet$ \textbf{Concat}: This method directly concatenates the retrieval results and the question for evaluation.

$\bullet$ \textbf{Prompt}: This method uses a manually-crafted prompt to indicate the use of retrieval information (details are provided in Appendix).
% \footnote{LLaMA-2: \url{https://huggingface.co/meta-llama/Llama-2-7b-chat-hf}, Mistral: \url{https://huggingface.co/mistralai/Mistral-7B-Instruct-v0.1}.} 

$\bullet$ \textbf{Prefix-tuning}~\cite{prefix-tuning}: This method uses prefix-tuning to fine-tune the backbone models. To make a fair comparison with our method, we add 50 prefix tokens for training.

$\bullet$ \textbf{LoRA}~\cite{lora}: This method uses LoRA to fine-tune the backbone models. We use the hyperparameters suggested by the LLaMA's official guidance.\footnote{LLaMA Recipes, \url{https://github.com/meta-llama/llama-recipes/blob/main/src/llama_recipes/configs/peft.py}} To further validates the effectiveness of our SPRING on models that have already been optimized for RAG, we also train our method based on the LoRA checkpoint, and denote this variant as SPRING$^+$.

% The implementation details are provided in Appendix.

\subsection{Implementation Details}\label{app:implementation}
We use PyTorch~\cite{pytorch} and Huggingface Accelerate library to implement all methods. The learning rate is set as $1\text{e-}4$ with a warm-up ratio of $0.1$. All methods are trained for three epochs, with a training batch size of $256$. We use eight NVIDIA A800 GPUs for training. Training our \ours{} for Mistral-7b models consumes around $2.2$ hours per epoch. The embeddings of the virtual tokens are initialized by the embeddings of the prompt: ``According to the previous relevant passages, please answer the following question. Only return the answer without any other words.'' Following the settings of prefix-tuning, if the number of tokens required exceeds those available in the prompt, the prompt is repeated to complete the initialization; if fewer are needed, the prompt is truncated accordingly. Additionally, we experiment with random initialization of tokens but observe that its performance is slightly worse than that achieved through prompt-based initialization. Our code is available at \url{https://github.com/DaoD/SPRING}.

\subsection{Experimental Results}
We fine-tune the prefix-tuning, LoRA, and \ours{} methods on RAG tasks, and then evaluate their performance in scenarios both with (RAG) and without (non-RAG) retrieval. For \ours{}, we use $k=50$ virtual tokens for inference by default, and the impact of token quantity $k$ is discussed in later. The experimental results are shown in Table~\ref{tab:in-domain}. To save space, we only show the results based on \texttt{Mistral-7b-instruct}, and other results are provided in Appendix. 

% @todo: add prompt's effectiveness
We can observe: 
(1)~It is evident that \ours{} significantly improves the RAG performance of the original LLM with manually-crafted prompts (the average EM and F1 scores are improved by $43.4\%$ and $17.3\%$, respectively). It outperforms LoRA on certain datasets, such as TriviaQA and CoQA. Given that \ours{} involves only 0.2M trainable parameters, these results demonstrate its remarkable efficiency and effectiveness. 
(2)~While LoRA achieves slightly better performance on some datasets, it adjusts the LLMs' original parameters, which adversely impact their performance in non-RAG scenarios—a significant drop has been observed, even far worse than the original models. This challenge also extends to other general generation tasks, which will be discussed in the next section. 
(3)~In non-RAG evaluation, only \ours{} and \ours{}$^+$ demonstrate better performance than the Prompt method. This indicates that even in the absence of retrieved results, adding virtual tokens is still beneficial. We speculate that beyond simply utilizing retrieved results, virtual tokens can help the LLM understand the task goal and format (\eg, the task is question-answering rather than text continuation).
(4)~Based on the LoRA's checkpoint, SPRING$^+$ achieves the best performance on most datasets. Additionally, all backbone models show improvements with \ours{} (see Appendix). These findings verify the versatility and flexibility of our approach, confirming its suitability for enhancing various LLMs in RAG scenarios.
(5)~Using manually-crafted prompts is effective for improving LLMs' performance on RAG tasks. However, this improvement is limited as no training is involved.
(6)~\ours{} achieves robust performance on the held-out datasets, validating the good generalizability of our method. 
(7)~Interestingly, prefix-tuning cannot perform well for RAG, highlighting that the insertion position of the virtual tokens in \ours{} is both reasonable and effective. We provide more analysis in Appendix.

% The evaluation results are shown in Figure~\ref{}. We can observe: (1) On the PopQA dataset, both \ours{} and LoRA demonstrate strong performance; however \ours{} behaves slightly better. This suggests better generalizability of our method. (2) In the non-RAG scenarios, while the the test datasets have been learned during training, the training process involves retrieved results. We see a significant performance drop in LoRA. This highlights a potential limitation of fine-tuning LLMs specifically for RAG tasks, as it may impair the models' general generation capabilities. (3) The performance drop of LoRA is more significant in other non-RAG tasks, implying that it negatively affects the LLMs' reasoning abilities and knowledge memorization. This further verifies the advantage of \ours{}, which effectively preserves the general capabilities of LLMs across various tasks.

\subsection{Further Analysis}
We further conduct a series of experiments to investigate the impact of different settings in \ours{}. All the following experiments are conducted based on fine-tuning the \texttt{Mistral-7b-instruct} model. 

\begin{table}[t]
    \centering
    \small
    \begin{tabular}{lcccc}
    \toprule
       Dataset & $n$-shot & LoRA & \ours{} & Diff \\
    \midrule
        BoolQ & 0 & 79.30 & \textbf{82.97} & 3.67 \\
        CommonsenseQA & 0 & 55.45 & \textbf{63.80} & 8.35 \\
        CommonsenseQA & 4 & 59.87 & \textbf{67.07} & 7.20 \\
        GSM8K & 8 & 17.33 & \textbf{31.89} & 14.56 \\
        MMLU & 0 & 51.30 & \textbf{53.62} & 2.32 \\
        MMLU & 5 & 48.76 & \textbf{54.96} & 6.20 \\
    \bottomrule
    \end{tabular}
    \caption{Performance comparison on other datasets.}
    %\vspace{-5px}
    \label{tab:non-rag}
\end{table}

\subsubsection{Performance on Other Tasks}
To examine the impact of different fine-tuning methods on the inherent capabilities of LLMs, we evaluate the performance of models fine-tuned by LoRA and \ours{} on several other (non-RAG) tasks. These tasks are commonly used to evaluate LLMs' reasoning, mathematical abilities, and world knowledge, including BoolQ~\cite{boolq}, CommonsenseQA~\cite{commonsenseqa}, GSM8K~\cite{gsm8k}, and MMLU~\cite{mmlu}. The experimental results are shown in Table~\ref{tab:non-rag}.\footnote{We notice that our results are quite different from those officially reported, which we attribute to the impact of different prompts. Since official prompts for testing are unavailable, we provide the prompts we used in Appendix to facilitate reproducibility.} From the results, we can observe: (1) Thanks to the plug-and-play design of our method, \ours{} can revert to to the original LLMs by not using virtual tokens. Therefore, it successfully preserves the original capabilities of the LLMs. In contrast, LoRA, which adjusts the model’s parameters for RAG tasks, inevitably compromises the model’s performance on other tasks. (2) A noticeable decline is observed in the few-shot evaluation, reflecting a decrease in the in-context learning abilities of LLMs. This decline may stem from the fact that RAG fine-tuning does not incorporate in-context learning capabilities. Besides, fine-tuning for RAG tasks may lead the model to overfit to specific task formats (prompts), thereby impairing its general generation abilities (more empirical studies are detailed in Appendix).

\begin{figure}[t]
    \centering
    \includegraphics[width=.9\linewidth]{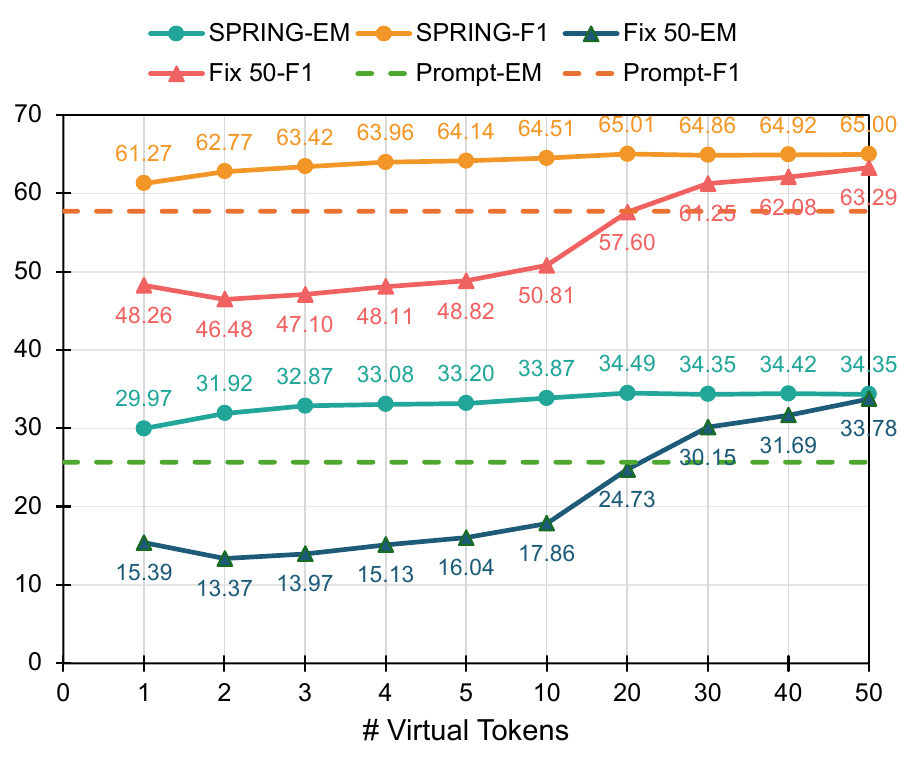}
    \caption{Average performance on nine QA datasets with various numbers of virtual tokens.}
    %\vspace{-5px}
    \label{fig:tokens}
\end{figure}

% @todo: fix-50 analysis
\subsubsection{Impact of Token Quantity}\label{sec:token_num}
In \ours{}, we design a scalable training approach that enables to use arbitrary numbers of virtual tokens in inference. To validate its effectiveness, we test the performance of our method with various numbers of virtual tokens and compare it with a variant model trained with a fixed number of tokens ($k=50$). The experimental results are illustrated in Figure~\ref{fig:tokens}. In general, we can observe that the performance of \ours{} increases with more virtual tokens used. Surprisingly, \ours{} can significantly enhance LLMs' performance in RAG scenarios with just a single token, which is very encouraging.\footnote{This varies across different LLMs (see Appendix).} In comparison, training with a fixed number of tokens limits the flexibility of \ours{}, making it can only be used with the same number of tokens in inference (\ie, $k=50$). 

% \begin{figure}[t]
%     \centering
%     \includegraphics[width=\linewidth]{figures/retriever.pdf}
%     \caption{Average performance on nine QA datasets with different retrievers.}
%     \label{fig:retrievers}
% \end{figure}

\begin{table}[t]
    \centering
    \small
    \begin{tabular}{lcccc}
    \toprule
        & \multicolumn{2}{c}{Prompt} & \multicolumn{2}{c}{\ours{}} \\
    \cmidrule(r){2-3} \cmidrule(l){4-5}
        Retriever & EM & F1 & EM & F1 \\
    \midrule
        BM25 & 21.23 & 54.94 & 30.94 & 62.73 \\
        BGE-base & 23.07 & 56.12 & 31.81 & 63.46 \\
        E5-base & 24.38 & 56.84 & 33.34 & 64.49 \\
        E5-large & 25.66 & 57.70 & \textbf{34.35} & \textbf{65.00} \\
    \midrule
        Average & 23.58 & 56.40 & 32.61 & 63.92 \\
        Variance & 2.69 & 1.02 & 1.75 & 0.78 \\
    \bottomrule
    \end{tabular}
    \caption{Average performance on nine QA datasets with different retrievers.}
    %\vspace{-5px}
    \label{tab:retrievers}
\end{table}

% \begin{figure*}[t]
%     \centering
%     \begin{subfigure}[b]{0.3\linewidth}
%         \centering
%         \includegraphics[width=\linewidth]{figures/token_num.pdf}
%         \caption{Average performance on nine QA datasets with various numbers of virtual tokens.}
%         \label{fig:tokens}
%     \end{subfigure}
%     \begin{subfigure}[b]{0.3\linewidth}
%         \centering
%         \includegraphics[width=\linewidth]{figures/retriever.pdf}
%         \caption{Average performance on nine QA datasets with various numbers of virtual tokens.}
%         \label{fig:tokens}
%     \end{subfigure}
%     \begin{subfigure}[b]{0.3\linewidth}
%         \centering
%         \includegraphics[width=\linewidth]{figures/retrieval_num.pdf}
%         \caption{Average performance on nine QA datasets with various numbers of virtual tokens.}
%         \label{fig:tokens}
%     \end{subfigure}
%     \hfill
%     \caption{Caption}
%     \label{fig:enter-label}
% \end{figure*}

\subsubsection{Effects of Different Retrievers}\label{sec:ret}
In our experiments, \ours{} is fine-tuned using passages retrieved by \ttt{E5-large}. To investigate its effectiveness with other retrievers, we conduct an experiment by testing its performance with passages retrieved by \ttt{BM25}, \ttt{BGE-base}, and \ttt{E5-large}. The results are presented in Table~\ref{tab:retrievers}. First, \ours{} achieves consistent improvement over the original model using manually crafted prompt, thereby confirming the generalizability of our approach. Second, compared to the original model, the performance gap (variance) among different retrievers becomes smaller, highlighting \ours{}'s robustness to variations in retrievers. Finally, even fine-tuned with a superior retriever (\ie, \ttt{E5-large}), \ours{} maintains strong performance well with less effective retrievers (such as \ttt{BM25}). This indicates that our method can effectively adapt to varying quality of retrieved results. Hence, there is no necessity to retrain the virtual tokens with each update of retrievers in practical applications, significantly enhancing its applicability.

\begin{figure}[t]
    \centering
    \includegraphics[width=.9\linewidth]{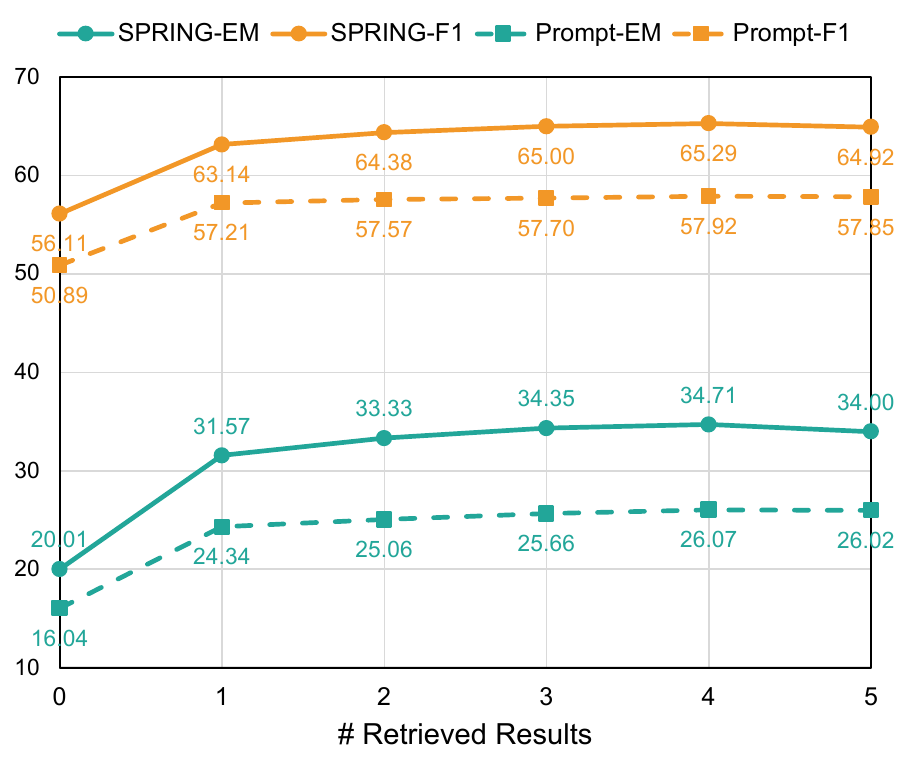}
    \caption{Average performance on nine QA datasets with different number of retrieved passages.}
    %\vspace{-5px}
    \label{fig:retrieval_num}
\end{figure}

\subsubsection{Influence of Retrieved Passages}
During the fine-tuning of \ours{}, we construct training samples by randomly selecting the top-$m$ ($m\in[1,5]$) retrieved passages. This aims to enhance \ours{}’s adaptability by ensuring it can operate effectively with varying numbers of retrieved passages in real-world scenarios. To evaluate the effect of this training strategy, we test the \ours{}'s performance across a range from zero to five passages. Figure~\ref{fig:retrieval_num} illustrates the results. We can find that \ours{}'s performance gradually improves as more retrieved passages are used ($m=0\rightarrow 4$), suggesting that more retrieved passages contribute valuable knowledge for question answering. However,  the performance peaks at four passages and declines when more passages are added. This decrease could be attributed to noise accumulation within the retrieved knowledge, a phenomenon also reported in recent studies~\cite{DBLP:journals/corr/abs-2310-01558,DBLP:journals/corr/abs-2404-03302}. Despite this, the use of retrieved passages still results in performance gains compared to scenarios without retrieval ($m=0$), highlighting again the benefits of RAG.

\begin{table}[t]
    \centering
    \small
    \setlength{\tabcolsep}{1.8mm}{
    \begin{tabular}{lcccccc}
    \toprule
        Training $\rightarrow$ & \multicolumn{2}{c}{TQA} & \multicolumn{2}{c}{NQ} & \multicolumn{2}{c}{Mix} \\
    \cmidrule(r){2-3} \cmidrule(lr){4-5} \cmidrule(l){6-7} 
        Test $\downarrow$ & EM & F1 & EM & F1 & EM & F1 \\
    \midrule
        TQA & 62.80 & 84.65 & 65.51 & 84.73 & \textbf{65.71} & \textbf{85.26} \\
        NQ &  32.19 & 64.98 & \textbf{45.26} & \textbf{72.68} & 42.35 & 70.73 \\
        HQA & 22.97 & 56.95 & 28.11 & 59.45 & \textbf{35.26} & \textbf{65.44}\\
        SQuAD & 25.24 & 61.39 & 30.81 & 64.32 & \textbf{33.67} & \textbf{66.99} \\
        WebQ & 26.03 & 62.40 & \textbf{34.13} & \textbf{67.30} & 31.84 & 64.78\\
        2Wiki & 17.32 & 54.41 & 25.43 & 58.21 & \textbf{31.80} & \textbf{62.03} \\
        CoQA & 6.05 & 36.61 & 7.62 & 38.54 & \textbf{13.28} & \textbf{42.41} \\
        MARCO & 3.26 & 31.88 & 3.88 & 33.33 & \textbf{6.57} & \textbf{53.48} \\
        PopQA & 44.07 & 72.50 & 48.28 & 74.07 & \textbf{48.71} & \textbf{73.90} \\
    \midrule
        Average & 26.66 & 58.42 & 32.11 & 61.40 & \textbf{34.35} & \textbf{65.00} \\
    \bottomrule
    \end{tabular}
    }
    \caption{Performance comparison between training on a specific dataset or a mixture of all datasets.}
    %\vspace{-5px}
    \label{tab:cross}
\end{table}

\subsubsection{Cross-Dataset Generalizability} 
Inspired by previous studies in multi-task learning~\cite{t5,unifiedqa}, we mix eight QA datasets for training as they require similar LLM capabilities (\eg, reasoning). To study the impact of this strategy, we conduct experiments by training \ours{} on each dataset individually and then testing its performance on the others. Table~\ref{tab:cross} shows partial results, and more results are available at Appendix. As indicated, training on a mixed dataset generally enhances performance on most datasets, thereby validating the benefits of multi-task learning. While training on a single dataset, such as NQ, may yield superior results on its specific test set, such improvements often fail to generalize to other datasets. Notably, training solely on NQ may negatively impact performance on MS MARCO, where the original LLM using a prompt could outperform it. These findings inspire us to carefully consider the interaction between different datasets when applying our method in future applications.

% \subsubsection{Impact of Token Position} (prefix vs stem vs suffix)

\section{Conclusion}
In this paper, we introduced scalable and pluggable virtual tokens for retrieval-augmented large language models. Our method, \ours{}, serves as a parameter-efficient fine-tuning approach that significantly enhances RAG performance with the addition of only $0.2$M trainable parameters. More importantly, the plug-and-play nature of our approach successfully preserves the performance of LLMs on non-RAG tasks, while its scalable training strategy broadens the method’s applicational flexibility. Through extensive experiments across various datasets, we have demonstrated the effectiveness, generalizability, flexibility, and high efficiency of our method. We believe that our research will foster further integration of information retrieval and LLMs, and advance the development of other parameter-efficient fine-tuning technologies for LLMs.

\section{Acknowledgment}
This work was supported by National Natural Science Foundation of China (Grant No. 62402497 and 62272467), Beijing Natural Science Foundation L233008, and Beijing Municipal Science and Technology Project  No. Z231100010323009. The work was partially done at the Engineering Research Center of Next-Generation Intelligent Search and Recommendation, MOE.

\clearpage
\begin{figure}[t]
    \centering
    \includegraphics[width=\linewidth]{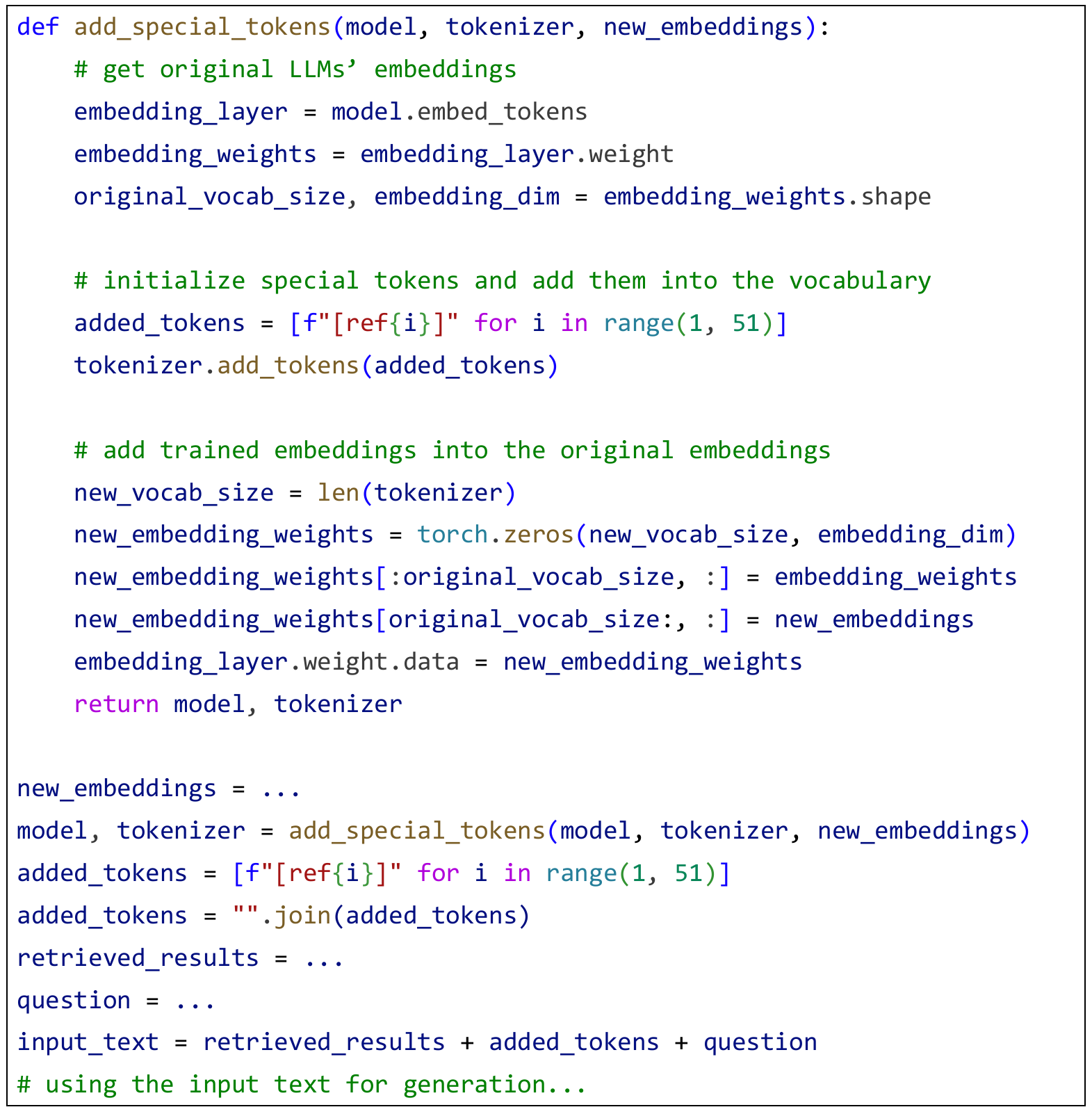}
    \caption{An example code snippet of using our \ours{} in practice.}
    \label{fig:code}
\end{figure}

\section{Example Code for Usage}\label{app:code}
We provide an example of how to use our \ours{} in practice. During the initialization phase, \ours{} merges the trained embeddings with the original embeddings of the LLM, and assigns these new embeddings to newly added special tokens. These special tokens are then inserted between the retrieved results and the user input for generation.

\begin{table*}[t]
    \centering
    \small
    \begin{tabular}{lrrrr}
    \toprule
        \textbf{QA Dataset} & \textbf{\# Training} & \textbf{\# Test} & \textbf{Average Input Length} & \textbf{Average Output Length} \\
    \midrule
        Trivia QA & 78,785 & 11,313 & 20.93 & 5.16\\
        Natural Questions & 79,168 & 3,610 & 11.31 & 4.66 \\
        Hotpot QA & 90,447 & 7,405 & 26.13 & 4.38 \\
        SQuAD 1.1 & 87,599 & 10,570 & 13.89 & 5.56 \\
        Web Questions & 3,778 & 2,032 & 10.04 & 4.10 \\
        2WikiMultiHopQA & 15,000 & 12,576 & 19.33 & 4.10 \\
        CoQA & 7,199 & 500 & 7.89 & 3.89\\
        MS MARCO & 80,000 & 55,636 & 7.76 & 20.75\\
        PopQA & - & 14,267 & 9.24 & 3.48 \\
        BoolQ & - & 3,270 & 10.70 & 1.00 \\
        CommonsenseQA & - & 1,221 & 15.91 & 1.00\\
        GSM8K & - & 1,319 & 64.90 & 3.29 \\
        MMLU & - & 14,042 & 63.91 & 9.62 \\
    \midrule
        \textbf{Retrieval Set} & \multicolumn{2}{r}{\textbf{\# Passages}} & \multicolumn{2}{r}{\textbf{Average Length}} \\
    \midrule
        Wikipedia Passages & \multicolumn{2}{r}{21,015,324} & \multicolumn{2}{r}{169.57} \\ %169.565
        MS MARCO Passages & \multicolumn{2}{r}{8,841,823} & \multicolumn{2}{r}{95.73} \\%95.729
    \bottomrule
    \end{tabular}
    \caption{Statistics of all datsaets. The input/output length is the number of tokens calculated by Mistral tokenizer.}
    \label{tab:statistics}
\end{table*}

\section{Datasets}\label{app:dataset}
We use the following datasets in our experiments, and their statistics are shown in Table~\ref{tab:statistics}.

(1) \textbf{TriviaQA}~\cite{triviaqa} is a reading comprehension dataset including question-answering pairs created by trivia enthusiasts, accompanied by independently gathered evidence documents. The questions are relatively complex and compositional, which require advanced reasoning to derive answers. This dataset is licensed under Apache 2.0 License.

(2) \textbf{Natural Questions}~\cite{nq} is a question-answering dataset. The questions are collected from the Google search engine, with answers annotated based on the Wikipedia pages. All questions are from real users. This dataset is licensed under Apache 2.0 License.

(3) \textbf{HotpotQA}~\cite{HotpotQA} is a question-answering dataset with natural and multi-hop questions. All question-answer pairs are Wikipedia-based, where the questions are diverse and include factoid comparison questions. This dataset is licensed under CC BY-SA 4.0 License.

(4) \textbf{SQuAD 1.1}~\cite{squad} is a reading comprehension dataset derived from Wikipedia articles. The questions are posed by crowd workers while the answers are text segments from the corresponding passages. This dataset is licensed under CC BY-SA 4.0 License.

(5) \textbf{Web Questions}~\cite{web_questions} is a question-answering dataset using Freebase as the knowledge base. The questions are collected via the Google Suggest API, and the answers are defined as Freebase entities. This dataset is licensed under CC BY 4.0 License.

(6) \textbf{2WikiMultiHopQA}~\cite{2wikimultihopqa} is a multi-hop question-answering dataset designed to test reasoning and inference skills. It introduces evidence information containing a reasoning path to ensure the quality and multi-hop nature of the questions. This dataset is licensed under Apache 2.0 License.

(7) \textbf{CoQA}~\cite{coqa} is a conversational question-answering dataset, which is obtained from human conversations about text passages from various domains. To keep the completeness of the question, we only use the first-round questions and their corresponding answers. This dataset contains data from several sources, under the license of CC BY-SA 4.0, MSR-LA, Apache, and RACE-specific License.

(8) \textbf{MS MARCO}~\cite{ms_marco} is a question-answering dataset featuring real Bing questions and human-generated answers. Since MS MARCO is extremely large, to balance the data amount in each dataset, we randomly sample 80,000 examples from the training set for fine-tuning. Besides, we also filter out questions without answers in the test set for evaluation. This dataset is licensed under the MIT License.

(9) \textbf{PopQA}~\cite{popqa} is an open-domain QA dataset with fine-grained and long-tail Wikidata entities, Wikipedia page views, and the information of relationship type. Since LLMs often struggle with less popular factual knowledge, retrieval augmentation is particularly beneficial for this dataset. This dataset is licensed under the MIT License.

(10) \textbf{BoolQ}~\cite{boolq} is a reading comprehension dataset, where each sample consists of a question, a passage excerpt, and a yes/no answer. The questions are often complex and require difficult entailment-like inference to solve. Therefore, it is often used to evaluate LLMs' text comprehension capability. This dataset is licensed under CC BY-SA 3.0 License.

(11) \textbf{CommonsenseQA}~\cite{commonsenseqa} is a commonsense question-answering dataset with multiple-choice questions designed to distinguish between related concepts. It is commonly used for evaluating LLMs' commonsense reasoning ability. This dataset is licensed under the MIT License.

(12) \textbf{GSM8K}~\cite{gsm8k} is a dataset that contains high-quality linguistically diverse grade school math word problems to measure models' ability of multi-step mathematical reasoning. This dataset is licensed under the MIT License.

(13) \textbf{MMLU}~\cite{mmlu} is a comprehensive aggregated benchmark. It contains 57 tasks to measure LLMs' knowledge across different subjects. This dataset is licensed under the MIT License.

\begin{figure}[t!]
    \centering
    \includegraphics[width=\linewidth]{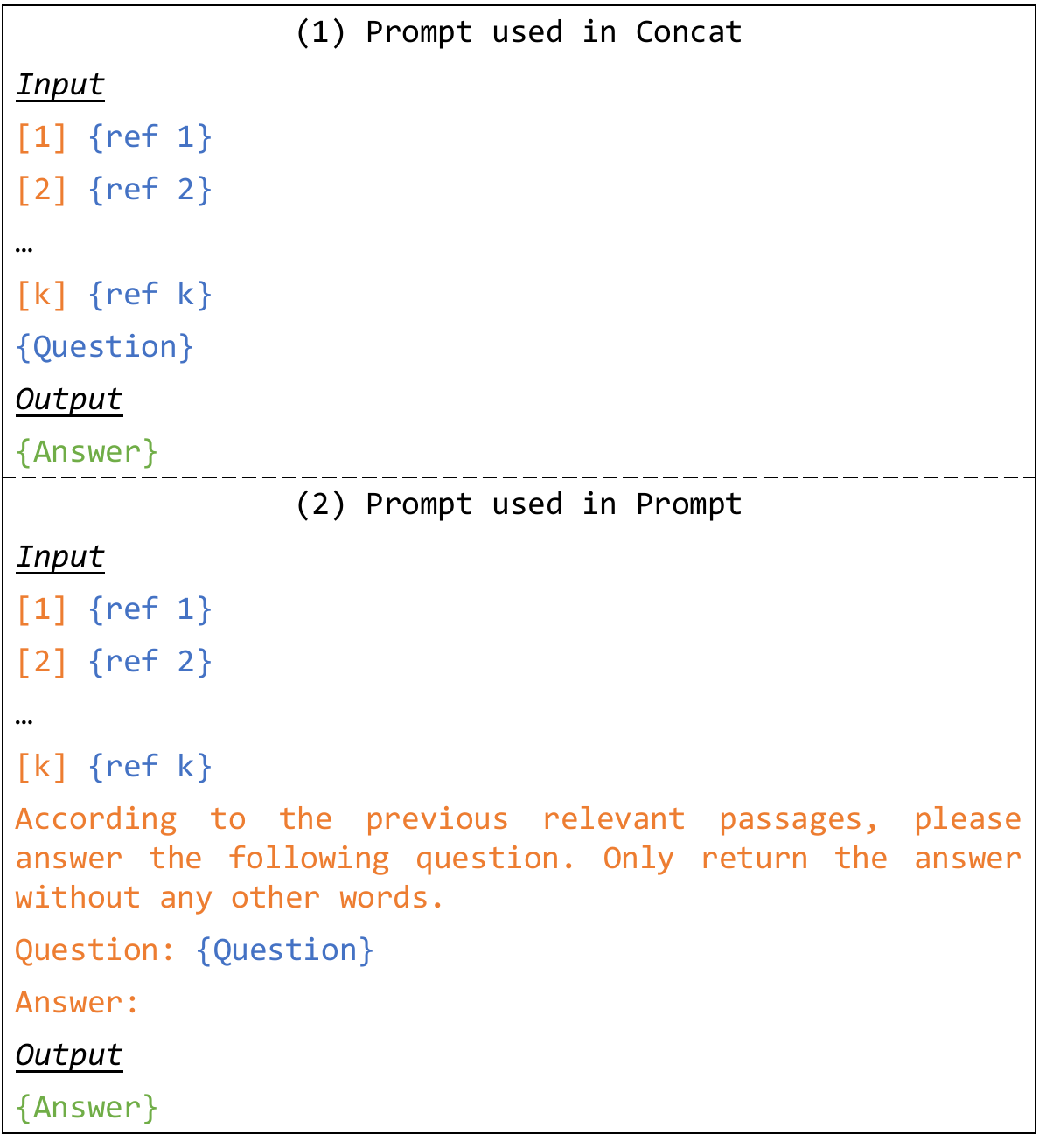}
    \caption{The prompt used in the baseline methods. ``\ttt{\{x\}}'' is the placeholder which will be replaced by the real data.}
    \label{fig:baseline_prompt}
\end{figure}

\begin{figure}[t!]
    \centering
    \includegraphics[width=\linewidth]{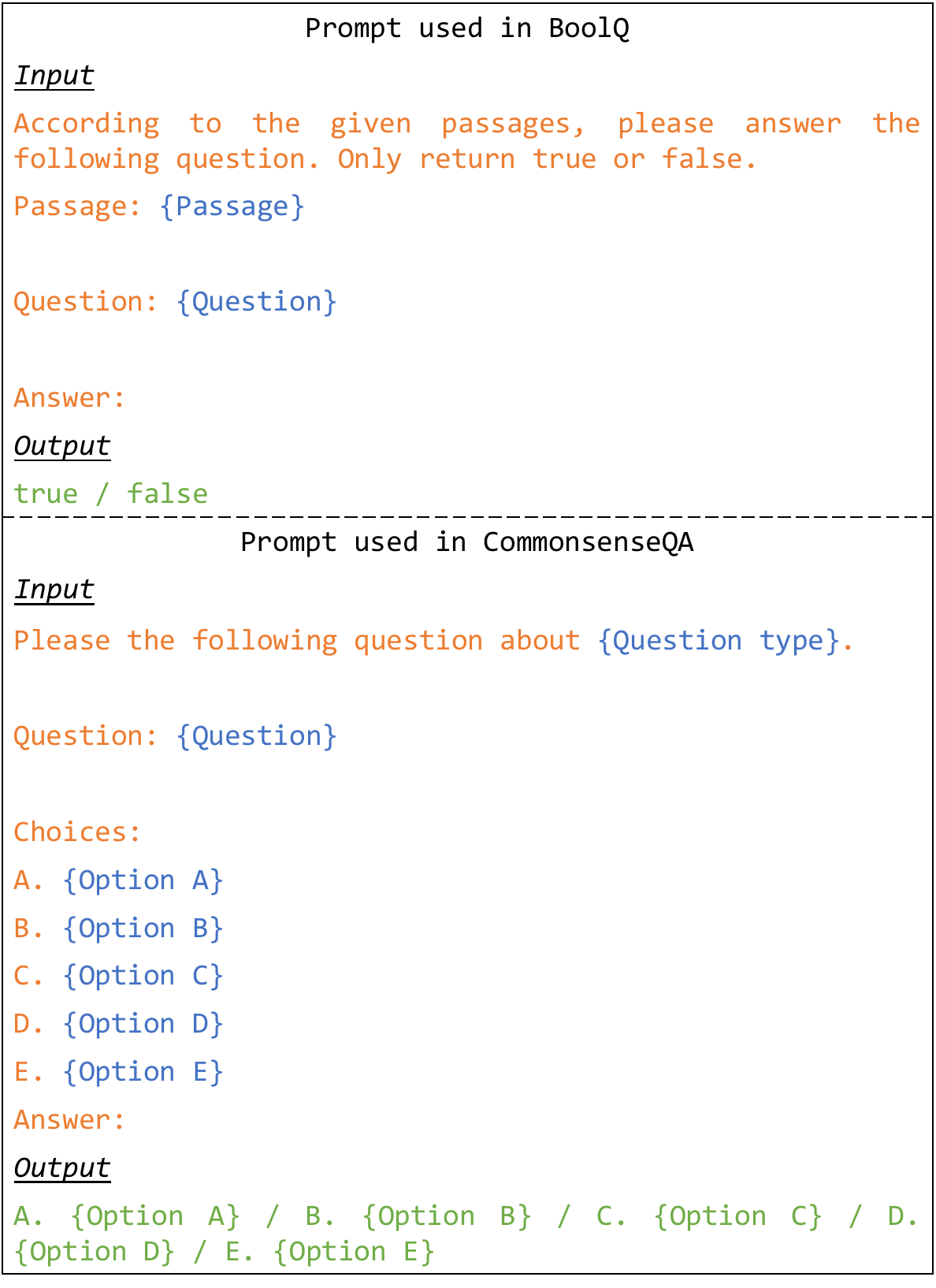}
    \caption{The prompt used in the BoolQ and CommonsenseQA datasets. ``\ttt{\{x\}}'' is the placeholder which will be replaced by the real data.}
    \label{fig:boolq_prompt}
\end{figure}

\begin{figure}[t!]
    \centering
    \includegraphics[width=\linewidth]{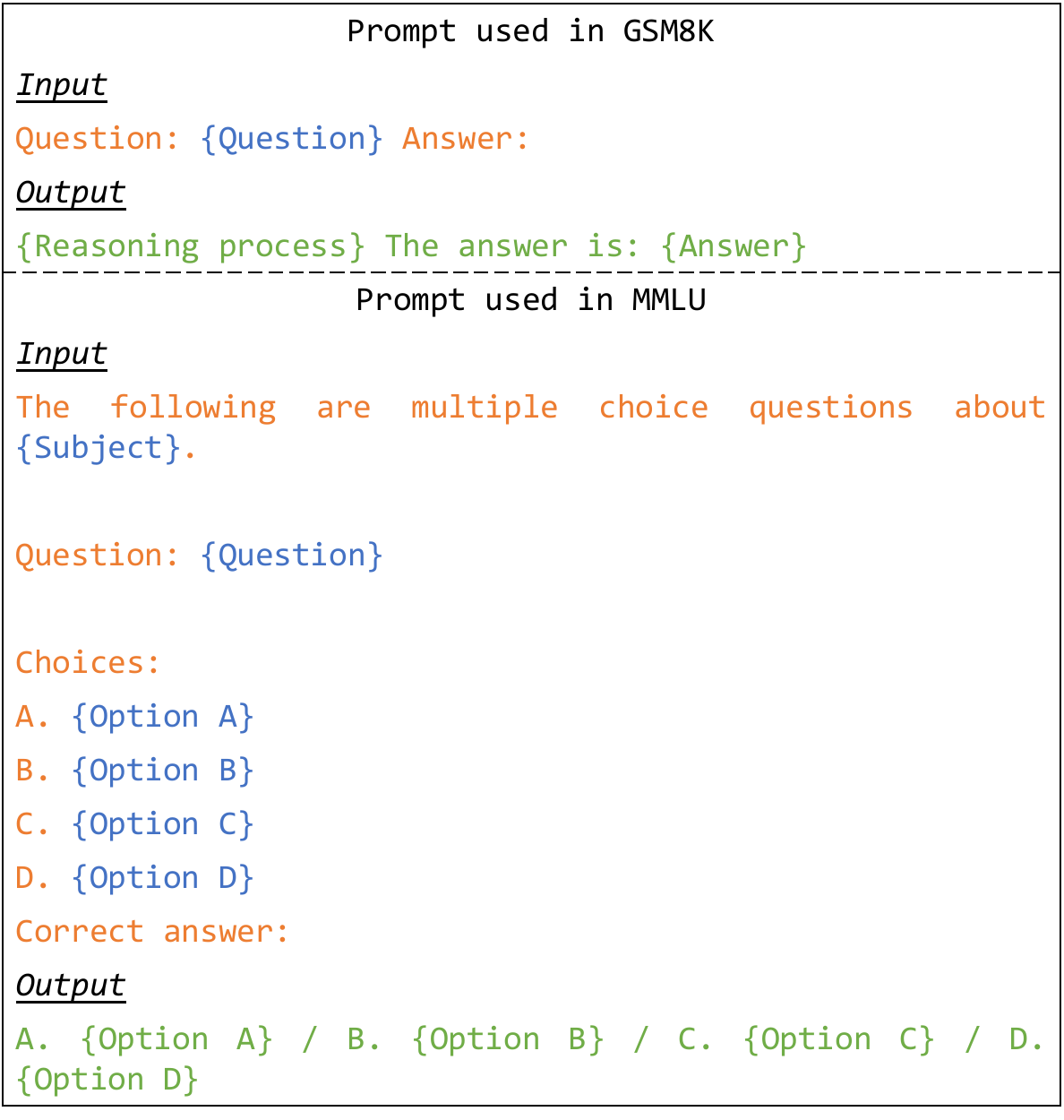}
    \caption{The prompt used in the GSM8K and MMLU datasets. ``\ttt{\{x\}}'' is the placeholder which will be replaced by the real data.}
    \label{fig:gsm8k_prompt}
\end{figure}

\section{Prompt}\label{app:prompt} 
Figure~\ref{fig:baseline_prompt} illustrates the prompts used in the baseline methods. In the Concat method, we simply concatenate the retrieval results and the question for evaluation. In the Prompt method, we use a manually-crafted prompt to indicate the incorporation of retrieval information.

Figure~\ref{fig:boolq_prompt} and Figure~\ref{fig:gsm8k_prompt} present the prompts used for the BoolQ, CommonsenseQA, GSM8K, and MMLU datasets. We follow the practices in \ttt{lm-evaluation-harness} and adopt different strategies for performing evaluation on these datasets.\footnote{\url{https://github.com/EleutherAI/lm-evaluation-harness}} Specifically, for the BoolQ dataset, we compare the generation probabilities of the token ``true'' and ``false'', and then we select the token with the higher probability as the predicted result. For CommonsenseQA and MMLU, we determine the answer by comparing the generation probabilities of each option with its content and choosing the option with the highest probability. For GSM8K, we provide LLMs with chain-of-thought demonstrations and extract the answer from the generated text. It is worth noting that the choice of prompt significantly affects the performance of LLMs, so we select the most effective prompt for the vanilla model and apply it across other methods for evaluation. We also observe that our results are worse than those officially reported, and we attribute the gap to the different prompts used for evaluation. Unfortunately, the official prompts are not released publicly, so we are limited to selecting the best-performing prompt from our experiments.

% \section{Implementation Details}\label{app:implementation}
% We use PyTorch~\cite{pytorch} and Huggingface Accelerate library to implement all methods. The learning rate is set as $1\text{e-}4$ with a warm-up ratio of $0.1$. All methods are trained for three epochs, with a training batch size of $256$. We use eight NVIDIA A800 GPUs for training. Training our \ours{} for Mistral-7b models consumes around $2.2$ hours per epoch. The embeddings of the virtual tokens are initialized by the embeddings of the prompt: ``According to the previous relevant passages, please answer the following question. Only return the answer without any other words.'' Following the settings of prefix-tuning, if the number of tokens required exceeds those available in the prompt, the prompt is repeated to complete the initialization; if fewer are needed, the prompt is truncated accordingly. Additionally, we experiment with random initialization of tokens but observe that its performance is slightly worse than that achieved through prompt-based initialization.

\begin{table*}[t]
    \centering
    \small
    \begin{tabular}{lcccccccc}
\toprule
    & \multicolumn{4}{c}{Mistral-7b-base} & \multicolumn{4}{c}{Mistral-7b-instruct} \\
\cmidrule(r){2-5}\cmidrule(l){6-9}
    & \multicolumn{2}{c}{Prompt} & \multicolumn{2}{c}{\ours{}} & \multicolumn{2}{c}{Prompt} & \multicolumn{2}{c}{\ours{}} \\
\cmidrule(r){2-3}\cmidrule(lr){4-5}\cmidrule(lr){6-7}\cmidrule(l){8-9}
    & EM & F1 & EM & F1 & EM & F1 & EM & F1 \\
\midrule
    TQA & 0.00 & 17.03 & \textbf{68.54} & \textbf{86.89} & 57.79 & 80.33 & \textbf{65.71} & \textbf{85.26} \\
    NQ & 0.08 & 7.47 & \textbf{46.90} & \textbf{73.28} & 28.99 & 58.72 & \textbf{42.35} & \textbf{70.73} \\
    HQA & 0.00 & 12.22 & \textbf{39.78} & \textbf{68.48} & 26.36 & 56.15 & \textbf{35.26} & \textbf{65.44} \\
    SQuAD & 0.01 & 12.04 & \textbf{37.13} & \textbf{69.02} & 23.92 & 57.66 & \textbf{33.67} & \textbf{66.99} \\
    WebQ & 0.00 & 11.61 & \textbf{35.25} & \textbf{66.44} & 17.53 & 52.10 & \textbf{31.84} & \textbf{64.78} \\
    2Wiki & 0.00 & 14.48 & \textbf{35.42} & \textbf{64.13} & 22.64 & 55.58 & \textbf{31.80} & \textbf{62.03} \\
    CoQA & 0.00 & 8.15 & \textbf{15.23} & \textbf{43.94} & 8.20 & 36.72 & \textbf{13.28} & \textbf{42.41} \\
    MS MARCO & 0.08 & 21.21 & \textbf{7.59} & \textbf{53.90} & 5.73 & \textbf{53.81} & \textbf{6.57} & 53.48 \\
    PopQA & 0.00 & 14.72 & \textbf{50.37} & \textbf{74.84} & 39.79 & 68.26 & \textbf{48.71} & \textbf{73.90} \\
\midrule
    Average & 0.02 & 13.21 & \textbf{37.36} & \textbf{66.77} & 25.66 & 57.70 & \textbf{34.35} & \textbf{65.00} \\
\bottomrule
    \end{tabular}
    \caption{Performance of \ours{} with Mistral-7b models.}
    \label{tab:mistral}
\end{table*}

\begin{table*}[t]
    \centering
    \small
    \begin{tabular}{lcccccccc}
\toprule
    & \multicolumn{4}{c}{LLaMA-2-7b-base} & \multicolumn{4}{c}{LLaMA-2-7b-chat} \\
\cmidrule(r){2-5}\cmidrule(l){6-9}
    & \multicolumn{2}{c}{Prompt} & \multicolumn{2}{c}{\ours{}} & \multicolumn{2}{c}{Prompt} & \multicolumn{2}{c}{\ours{}} \\
\cmidrule(r){2-3}\cmidrule(lr){4-5}\cmidrule(lr){6-7}\cmidrule(l){8-9}
    & EM & F1 & EM & F1 & EM & F1 & EM & F1 \\
\midrule
    TQA & 0.01 & 34.37 & \textbf{65.42} & \textbf{85.30} & 59.84 & 81.75 & \textbf{66.95} & \textbf{85.59} \\
    NQ & 0.00 & 22.31 & \textbf{43.48} & \textbf{71.04} & 32.46 & 62.10 & \textbf{42.13} & \textbf{70.57} \\
    HQA & 0.00 & 26.17 & \textbf{35.48} & \textbf{65.59} & 26.39 & 56.58 & \textbf{33.88} & \textbf{64.47} \\
    SQuAD & 0.00 & 26.49 & \textbf{35.10} & \textbf{67.34} & 24.57 & 59.40 & \textbf{33.36} & \textbf{66.60} \\
    WebQ & 0.00 & 23.76 & \textbf{31.20} & \textbf{63.73} & 20.85 & 57.22 & \textbf{31.10} & \textbf{64.31} \\
    2Wiki & 0.00 & 25.83 & \textbf{30.61} & \textbf{60.90} & 20.69 & 55.82 & \textbf{30.86} & \textbf{61.45} \\
    CoQA & 0.00 & 16.08 & \textbf{13.67} & \textbf{42.73} & 8.59 & 38.56 & \textbf{14.06} & \textbf{43.99} \\
    MS MARCO & 0.00 & 51.75 & \textbf{7.43} & \textbf{54.11} & 5.32 & 48.42 & \textbf{7.18} & \textbf{52.85} \\
    PopQA & 0.00 & 32.17 & \textbf{49.93} & \textbf{74.41} & 45.08 & 72.23 & \textbf{49.00} & \textbf{73.98} \\
\midrule
    Average & 0.00 & 28.77 & \textbf{34.70} & \textbf{65.02} & 27.09 & 59.12 & \textbf{34.28} & \textbf{64.87} \\
\bottomrule
    \end{tabular}
    \caption{Performance of \ours{} with LLaMA-2-7b models.}
    \label{tab:llama-7b}
\end{table*}

\begin{table*}[t]
    \centering
    \small
    \begin{tabular}{lcccccccc}
\toprule
    & \multicolumn{4}{c}{LLaMA-2-13b-base} & \multicolumn{4}{c}{LLaMA-2-13b-chat} \\
\cmidrule(r){2-5}\cmidrule(l){6-9}
    & \multicolumn{2}{c}{Prompt} & \multicolumn{2}{c}{\ours{}} & \multicolumn{2}{c}{Prompt} & \multicolumn{2}{c}{\ours{}} \\
\cmidrule(r){2-3}\cmidrule(lr){4-5}\cmidrule(lr){6-7}\cmidrule(l){8-9}
    & EM & F1 & EM & F1 & EM & F1 & EM & F1 \\
\midrule
    TQA & 0.26 & 30.77 & \textbf{66.95} & \textbf{86.48} & 63.54 & 82.54 & \textbf{68.52} & \textbf{86.62} \\
    NQ & 0.57 & 19.54 & \textbf{44.91} & \textbf{71.70} & 36.66 & 67.03 & \textbf{43.80} & \textbf{71.15} \\
    HQA & 0.07 & 19.72 & \textbf{37.24} & \textbf{66.81} & 29.27 & 59.36 & \textbf{36.23} & \textbf{66.11} \\
    SQuAD & 0.26 & 22.79 & \textbf{36.35} & \textbf{68.48} & 28.45 & 61.92 & \textbf{34.80} &\textbf{67.76} \\
    WebQ & 0.24 & 20.77 & \textbf{32.32} & \textbf{64.54} & 20.61 & 55.75 & \textbf{31.01} & \textbf{64.17} \\
    2Wiki & 0.10 & 22.63 & \textbf{33.32} & \textbf{62.85} & 28.01 & 59.20 & \textbf{32.32} & \textbf{62.43} \\
    CoQA & 0.39 & 14.85 & \textbf{14.84} & \textbf{44.00} & 9.96 & 39.27 & \textbf{14.26} & \textbf{44.49} \\
    MS MARCO & 0.57 & 47.30 & \textbf{7.36} & \textbf{53.71} & 6.24 & 47.68 & \textbf{7.51} & \textbf{52.94} \\
    PopQA & 0.13 & 24.06 & \textbf{50.08} & \textbf{74.72} & 42.49 & 69.67 & \textbf{50.29} & \textbf{74.84} \\
\midrule
    Average & 0.29 & 24.71 & \textbf{35.93} & \textbf{65.92} & 29.47 & 60.27 & \textbf{35.42} & \textbf{65.61} \\
\bottomrule
    \end{tabular}
    \caption{Performance of \ours{} with LLaMA-2-13b models.}
    \label{tab:llama-13b}
\end{table*}

\begin{table*}[t]
    \centering
    \small
    \begin{tabular}{lcccccccccc}
    \toprule
        Training $\rightarrow$ & \multicolumn{2}{c}{TQA} & \multicolumn{2}{c}{NQ} & \multicolumn{2}{c}{HQA} & \multicolumn{2}{c}{SQuAD} & \multicolumn{2}{c}{Mix} \\
    \cmidrule(r){2-3} \cmidrule(lr){4-5} \cmidrule(lr){6-7} \cmidrule(lr){8-9} \cmidrule(l){10-11} 
        Test $\downarrow$ & EM & F1 & EM & F1 & EM & F1 & EM & F1 & EM & F1 \\
    \midrule
        TQA & 62.80 & 84.65 & 65.51 & 84.73 & 66.42 & 85.26 & \textbf{67.48} & \textbf{85.76} & 65.71 & 85.26 \\
        NQ & 32.19 & 64.98 & \textbf{45.26} & \textbf{72.68} & 40.79 & 70.18 & 37.88 & 67.23 & 42.35 & 70.73 \\
        HQA & 22.97 & 56.95 & 28.11 & 59.45 & \textbf{36.30} & \textbf{66.33} & 30.43 & 61.46 & 35.26 & 65.44 \\
        SQuAD & 25.24 & 61.39 & 30.81 & 64.32 & 33.37 & 66.36 & \textbf{35.01} & \textbf{67.83} & 33.67 & 66.99 \\
        WebQ & 26.03 & 62.40 & \textbf{34.13} & \textbf{67.30} & 29.15 & 64.04 & 28.27 & 63.38 & 31.84 & 64.78 \\
        2Wiki & 17.32 & 54.41 & 25.43 & 58.21 & 30.42 & 61.62 & 25.48 & 58.26 & \textbf{31.80} & \textbf{62.03} \\
        CoQA & 6.05 & 36.61 & 7.62 & 38.54 & 12.50 & \textbf{43.12} & 7.62 & 37.74 & \textbf{13.28} & 42.41 \\
        MS MARCO & 3.26 & 31.88 & 3.88 & 33.33 & 5.18 & 36.12 & 3.89 & 45.80 & \textbf{6.57} & \textbf{53.48} \\
        PopQA & 44.07 & 72.50 & 48.28 & 74.07 & \textbf{48.80} & \textbf{74.32} & 48.40 & 73.83 & 48.71 & 73.90 \\
    \midrule
        Average & 26.66 & 58.42 & 32.11 & 61.40 & 33.66 & 63.04 & 31.61 & 62.37 & \textbf{34.35} & \textbf{65.00} \\
    \bottomrule
    \end{tabular}
    \caption{Full results of training on a specific dataset and a mixture of all datasets.}
    \label{tab:my_label}
\end{table*}

\section{Full Experimental Results}\label{app:results}
In the main body of our paper, we present the results using \ttt{Mistral-7b-instruct} as the backbone model. We also apply our method to \ttt{Mistral-7b-base}, \ttt{LLaMA-2-7b-chat}, \ttt{LLaMA-2-7b-base}, \ttt{LLaMA-2-13b-base}, and \ttt{LLaMA-2-13b-chat}. Their results are shown in Table~\ref{tab:mistral}, Table~\ref{tab:llama-7b}, and Table~\ref{tab:llama-13b}. We can observe that \ours{} can achieve consistent improvement for all models. Specifically, using prompt is very effective for instruction-tuned models~(\ie, Mistral-instruct and LLaMA-chat) but less so for base models. This is because the base model cannot understand the instructions to read the retrieved results and provide correct answers. Mistral series models perform slightly better than LLaMA series models of the same size. This is consistent with their officially reported benchmark performance. Furthermore, larger models tend to perform better than smaller models, and our \ours{} can bring improvement across models of different sizes. In future, we plan to extend our method to much larger models~(such as 30B or 70B) when more computing resources available.

\begin{table}[t]
    \centering
    \small
    \setlength{\tabcolsep}{1mm}{
    \begin{tabular}{lcccccc}
    \toprule
        & \multicolumn{2}{c}{$[T,R,Q]$} & \multicolumn{2}{c}{\ours{}} & \multicolumn{2}{c}{$[R,Q,T]$} \\
    \cmidrule(r){2-3}\cmidrule(lr){4-5}\cmidrule(l){6-7}
        & EM & F1 & EM & F1 & EM & F1 \\
    \midrule
        TQA & 11.74 & 59.97 & \textbf{65.71} & \textbf{85.26} & 64.90 & 84.65 \\
        NQ & 13.04 & 38.22 & 42.35 & 70.73 & \textbf{43.16} & \textbf{71.30} \\
        HQA & 5.79 & 42.59 & \textbf{35.26} & \textbf{65.44} & 34.00 & 64.27 \\
        SQuAD & 7.19 & 39.75 & 33.67 & \textbf{66.99} & \textbf{34.09} & 66.75 \\
        WebQ & 4.44 & 31.55 & \textbf{31.84} & 64.78 & 31.35 & \textbf{64.84} \\
        2Wiki & 4.38 & 41.82 & \textbf{31.80} & \textbf{62.03} & 30.64 & 61.01 \\
        CoQA & 1.56 & 20.02 & \textbf{13.28} & \textbf{42.41} & 12.30 & 41.18 \\
        MS MARCO & 0.60 & 50.56 & 6.57 & \textbf{53.48} & \textbf{6.94} & 49.40 \\
        PopQA & 10.02 & 44.54 & \textbf{48.71} & \textbf{73.90} & 46.55 & 72.33 \\
    \midrule
        Average & 6.53 & 41.00 & \textbf{34.35} & \textbf{65.00} & 33.77 & 63.97 \\
    \bottomrule
    \end{tabular}
    }
    \caption{Performance of different virtual token insertion positions. The best results are in bold.}
    \label{tab:position}
\end{table}

\section{Impact of Token Insertion Position}\label{app:position}
In \ours{}, virtual tokens~($T$) are strategically inserted between the retrieved results~($R$) and the question~($Q$). Generally, there are three possible insertion positions:
\begin{itemize}
    \item $[T, R, Q]$: The virtual tokens are added at the beginning of the sequence, akin to prefix-tuning. In this setup, due to the auto-regressive nature of LLMs, these tokens cannot access subsequent sequence information.
    \item $[R, T, Q]$: This is what we used in \ours{}. The virtual tokens can attend to the retrieved results, while positioning the question immediately next to the answer. 
    \item $[R, Q, T]$: The virtual tokens are added at the end of the sequence. They are able to attend to all preceding information, including both the retrieved results and the question.
\end{itemize}
We compare the performance of these three strategies. The results are shown in Table~\ref{tab:position}. It is evident that \ours{} achieves the best performance among the three strategies, which demonstrates the effectiveness of our token placement design. Placing virtual tokens at the beginning of the sequence~(prefix-tuning) is the worst strategy, because the added tokens cannot adapt based on the retrieved results or the question during fine-tuning. This result is different from those reported in the original prefix-tuning study~\cite{prefix-tuning}, which suggested that prefix-tuning can yield better performance. We attribute this discrepancy to the different targets of general generation tasks versus RAG tasks. In general generation, all contexts are crucial for predicting the next token, hence adding trainable tokens at the beginning can effectively influence the activation of contexts. In contrast, in RAG tasks, the retrieved information is supplementary to answering the user's question, so the virtual tokens are added to help LLMs understand and leverage the retrieved information rather than enhance the retrieved information itself. This speculation is further validated by the strong performance achieved by adding tokens at the end of the sequence, where they can also access the retrieved results. Among the three strategies, \ours{} is the most effective, not only allowing the virtual tokens to utilize the retrieved information optimally but also maintaining coherence between the question and the answer being generated.

\begin{table}[t!]
    \centering
    \small
    \begin{tabular}{p{.9\linewidth}}
    \toprule
         \textbf{Prompt 1}: \\
         According to the previous relevant passages, please answer the following question. Only return the answer without any other words.$\backslash$n \\
    \midrule
         \textbf{Prompt 2}: \\
         According to the previous relevant passages, please answer the following question.$\backslash$n \\
    \midrule
         \textbf{Prompt 3}: \\
         Answer the following question based on the provided passages.$\backslash$n \\
    \midrule
         \textbf{Prompt 4}: \\
         (None) \\
    \bottomrule
    \end{tabular}
    \caption{The prompts we used for exploring their impacts on inference.}
    \label{tab:prompt}
\end{table}

\begin{figure}[t!]
    \centering
    \includegraphics[width=\linewidth]{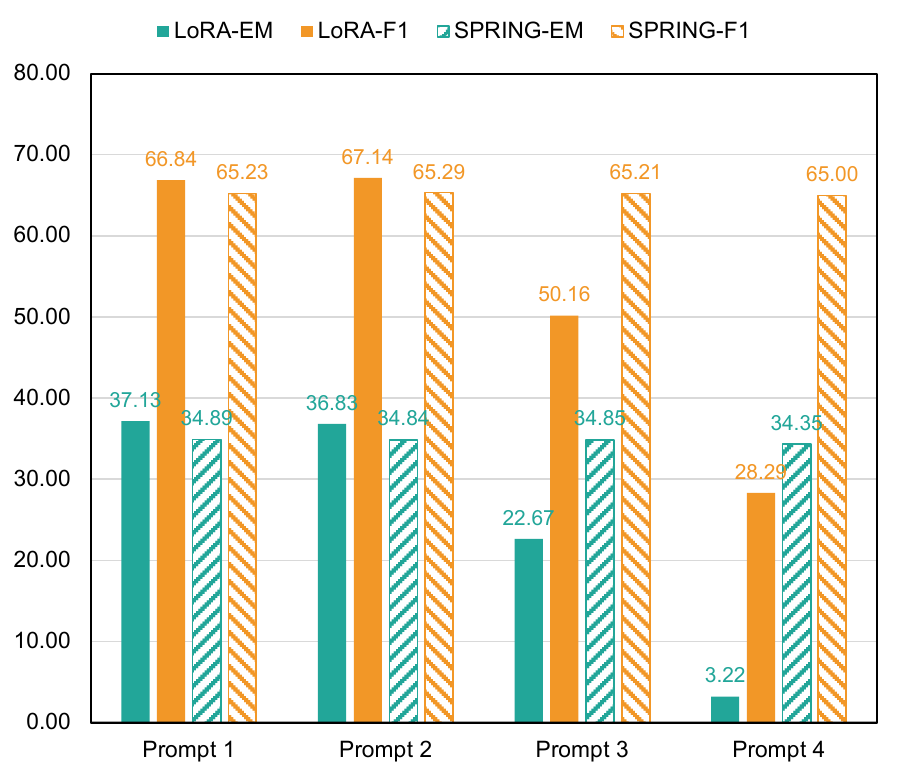}
    \caption{Performance under various prompts.}
    \label{fig:prompt}
\end{figure}

\section{Impact of Prompt after Fine-tuning}\label{app:test_prompt}
In our experiments, we notice that models fine-tuned by certain methods~(such as LoRA) are very sensitive to the prompts used for inference. To investigate this problem, we conduct an empirical study. We mainly focus on the impact of the prompt $P$ inserted between the retrieved results and the question, formatted as $[R; P; Q]$. We consider four different prompts listed in Table~\ref{tab:prompt}. Specifically, for LoRA, we use the first prompt during training, and test its performance with all prompts. For \ours{}, it does not need a prompt in training, which can be seen as using the fourth prompt~(none). We also evaluate its performance with the four prompts. 

The experimental results are shown in Figure~\ref{fig:prompt}. It clearly reflects the prompt dependency of models fine-tuned with LoRA. LoRA achieves the highest EM score of $37.13$ with the first prompt, which is used during its training. The performance remains robust with the second prompt, which is very similar to the first but omits the last sentence. However, the third prompt, while semantically similar to the first, involves different phrasing and significantly impacts LoRA's performance. Notably, in scenarios where no prompt is used~(the fourth prompt), LoRA's performance drops sharply, even worse than that of the original model. These results indicate that fine-tuning with LoRA may lead to overfitting problems. This may also be the reason why models tuned with LoRA cannot perform well on non-RAG tasks. In comparison, our \ours{} demonstrates a consistent performance across all prompts, demonstrating its robustness and generalizability. 

\begin{figure}[t]
    \centering
    \includegraphics[width=\linewidth]{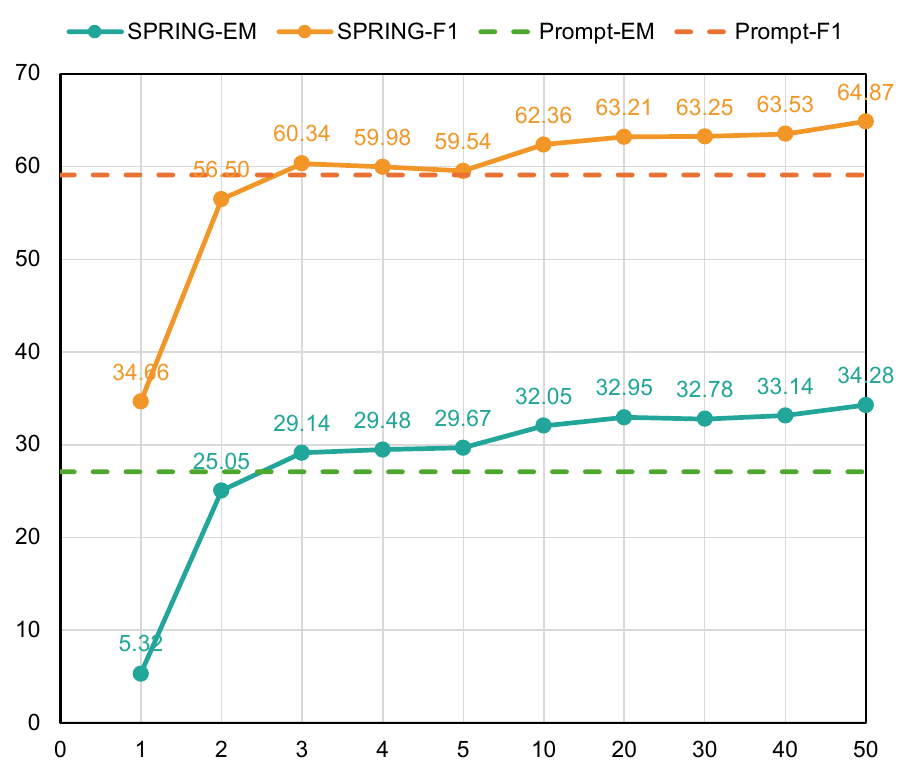}
    \caption{Average performance on nine QA datasets with various numbers of virtual tokens.}
    \label{fig:token_num_llama}
\end{figure}

\section{Impact of Token Quantity}\label{app:token_num}
In the main body of our paper, we explore the impact of token quantity based on the \texttt{Mistral-7b-instruct} model. Since this impact may vary across different models, we also extend our analysis to include the \texttt{LLaMA-2-7b-chat} model. The result is shown in Figure~\ref{fig:token_num_llama}. It can be seen that the general trend is similar to that based on the Mistral model, namely the performance of \ours{} is gradually improved as more virtual tokens are used. However, LLaMA requires at least three tokens to obtain improvement over the original model with prompts. This indicates that different models may require various numbers of virtual tokens to effectively learn and perform RAG tasks.

\begin{table*}[h!]
    \centering
    \small
    \begin{tabular}{lcccccc}
    \toprule
        Method & NQ & TriviaQA & HotpotQA & 2Wiki & PopQA & WebQA \\
        Metric & EM & EM & F1 & F1 & F1 & EM \\
    \midrule
        Naive Generation & 22.6 & 55.7 & 28.4 & 33.9 & 21.7 & 18.8 \\
        Standard RAG & 35.1 & 58.9 & 35.3 & 21.0 & 36.7 & 15.7 \\
        AAR~\cite{aar} & 30.1 & 56.8 & 33.4 & 19.8 & 36.1 & 16.1 \\
        LongLLMLingua~\cite{longllmlingua} & 32.2 & 59.2 & 37.5 & 25.0 & 38.7 & 17.5 \\
        RECOMP~\cite{recomp} & 33.1 & 56.4 & 37.5 & 32.4 & 39.9 & 20.2 \\
        Selective-Context~\cite{selective} & 30.5 & 55.6 & 34.4 & 18.5 & 33.5 & 17.3 \\
        Trace~\cite{trace} & 30.7 & 50.2 & 34.0 & 15.5 & 37.4 & 19.9 \\
        SuRe~\cite{sure} & 37.1 & 53.2 & 33.4 & 20.6 & 48.1 & 24.2\\
        REPLUG~\cite{replug} & 28.9 & 57.7 & 31.2 & 21.1 & 27.8 & 20.2\\
        SKR~\cite{skr} & 33.2 & 56.0 & 32.4 & 23.4 & 31.7 & 17.0\\
        Self-RAG~\cite{self-rag} & 36.4 & 38.2 & 29.6 & 25.1 & 32.7 & 21.9\\
        FLARE~\cite{flare} & 22.5 & 55.8 & 28.0 & 33.9 & 20.7 & 20.2\\
        ITRG~\cite{itrg} & 36.8 & 60.1 & 38.3 & 21.6 & 37.9 & 18.2\\
        IRCoT~\cite{ircot} & 33.3 & 56.9 & 41.5 & 32.4 & 45.6 & 20.7\\
    \midrule
        SPRING & \textbf{37.9} & \textbf{64.6} & \textbf{42.6} & \textbf{37.3} & \textbf{54.8} & \textbf{27.7}\\
    \bottomrule
    \end{tabular}
    \caption{Comparison with other RAG methods. All experimental results are cited from FlashRAG~\cite{flashrag}. Only \ours{} uses \texttt{LLaMA-2-7b} as the backbone model, while other methods use \texttt{LLaMA-3-8b}. Nevertheless, as the retrieval corpus and experimental settings used in FlashRAG are quite different from those used in our experiments, their results are also different from those reported in our paper.}
    \label{tab:other_rag}
\end{table*}

\section{Comparison with Other RAG Methods}
We further evaluate our \ours{} against other RAG methods. The experimental results, provided by FlashRAG~\cite{flashrag}, are shown in Table~\ref{tab:other_rag}. It is evident to see that \ours{} outperforms all other methods under the same settings. It is worth noting that only \ours{} uses \texttt{LLaMA-2-7b} as the backbone model, whereas other methods employ the more advanced \texttt{LLaMA-3-8b}. This clearly reflects the superiority of our method. Essentially, most existing methods rely primarily on complex prompt designs without fine-tuning, so they cannot achieve comparable performance with \ours{}. 

\end{document}